\documentclass{article}

\usepackage{times}
\usepackage{epsfig}
\usepackage{graphicx}
\usepackage{float}
\usepackage{wrapfig}

\usepackage{amsmath,amssymb}

\usepackage{bm,xspace}
\usepackage{comment}
\usepackage{verbatim}
\usepackage{multirow}
\usepackage{balance}
\usepackage{url}
\usepackage{booktabs}
\usepackage{etoolbox,siunitx}
\usepackage{calc}
\usepackage{pifont,hologo}
\usepackage{color}

\usepackage[super]{nth}
\usepackage{nicefrac}
\def\Re{\mathbb{R}}

\DeclareMathSymbol{@}{\mathord}{letters}{"3B}

\newcommand\norm[1]{\left\lVert#1\right\rVert}
\newcommand\tuple[1]{\left\langle#1\right\rangle}

\def\latex/{\LaTeX}
\def\bibtex/{\hologo{BibTeX}}

\usepackage[official]{eurosym}

\newcommand{\ie}{\emph{i.e.~}}
\newcommand{\eg}{\emph{e.g.~}}
\newcommand{\etc}{\emph{etc.\xspace}}

\newcommand{\vs}{\emph{vs.~}}

\newcommand{\figLabel}{Figure\xspace}

\newcommand{\secLabel}{Section\xspace}
\newcommand{\tblLabel}{Table\xspace}

\newcommand*{\mydprime}{^{\prime\prime}\mkern-1.2mu}
\newcommand*{\mytrprime}{^{\prime\prime\prime}\mkern-1.2mu}

\newcommand{\mysection}[1]{\vspace{0pt}\noindent\textbf{#1.}}

\usepackage{microtype}
\usepackage{graphicx}
\usepackage{subfigure}
\usepackage{booktabs} % for professional tables
\usepackage{multirow}
\usepackage{braket}
\usepackage{xcolor}

\usepackage{hyperref}

\usepackage[accepted]{icml2021}

\icmltitlerunning{Training Graph Neural Networks with 1000 Layers}

\begin{document}

\twocolumn[
\icmltitle{Training Graph Neural Networks with 1000 Layers}



\icmlsetsymbol{equal}{*}

\begin{icmlauthorlist}
\icmlauthor{Guohao Li}{intel,kaust}
\icmlauthor{Matthias M\"uller}{intel}
\icmlauthor{Bernard Ghanem}{kaust}
\icmlauthor{Vladlen Koltun}{intel}
\end{icmlauthorlist}

\icmlaffiliation{intel}{Intel Labs}
\icmlaffiliation{kaust}{King Abdullah University of Science and Technology}

\icmlcorrespondingauthor{Guohao Li}{guohao.li@kaust.edu.sa}

\icmlkeywords{Machine Learning, ICML, Graph Neural Networks}

\vskip 0.3in
]

\printAffiliationsAndNotice{}

\begin{abstract}
Deep graph neural networks (GNNs) have achieved excellent results on various tasks on increasingly large graph datasets with millions of nodes and edges. However, memory complexity has become a major obstacle when training deep GNNs for practical applications due to the immense number of nodes, edges, and intermediate activations. To improve the scalability of GNNs, prior works propose smart graph sampling or partitioning strategies to train GNNs with a smaller set of nodes or sub-graphs. In this work, we study 
reversible connections, group convolutions, weight tying, and equilibrium models to advance the memory and parameter efficiency of GNNs. 
We find that reversible connections in combination with deep network architectures enable the training of overparameterized GNNs that significantly outperform existing methods on multiple datasets. Our models RevGNN-Deep (1001 layers with 80 channels each) and RevGNN-Wide (448 layers with 224 channels each) were both trained on a single commodity GPU and achieve an ROC-AUC of 87.74 ± 0.13 and 88.24 ± 0.15 on the \emph{ogbn-proteins} dataset. To the best of our knowledge, RevGNN-Deep is the deepest GNN in the literature by one order of magnitude.
\end{abstract}

\section{Introduction}
\label{sec:intro}

\begin{figure}[h]
    \centering
    \includegraphics[trim=0cm 1cm 0cm 0cm, width=\columnwidth]{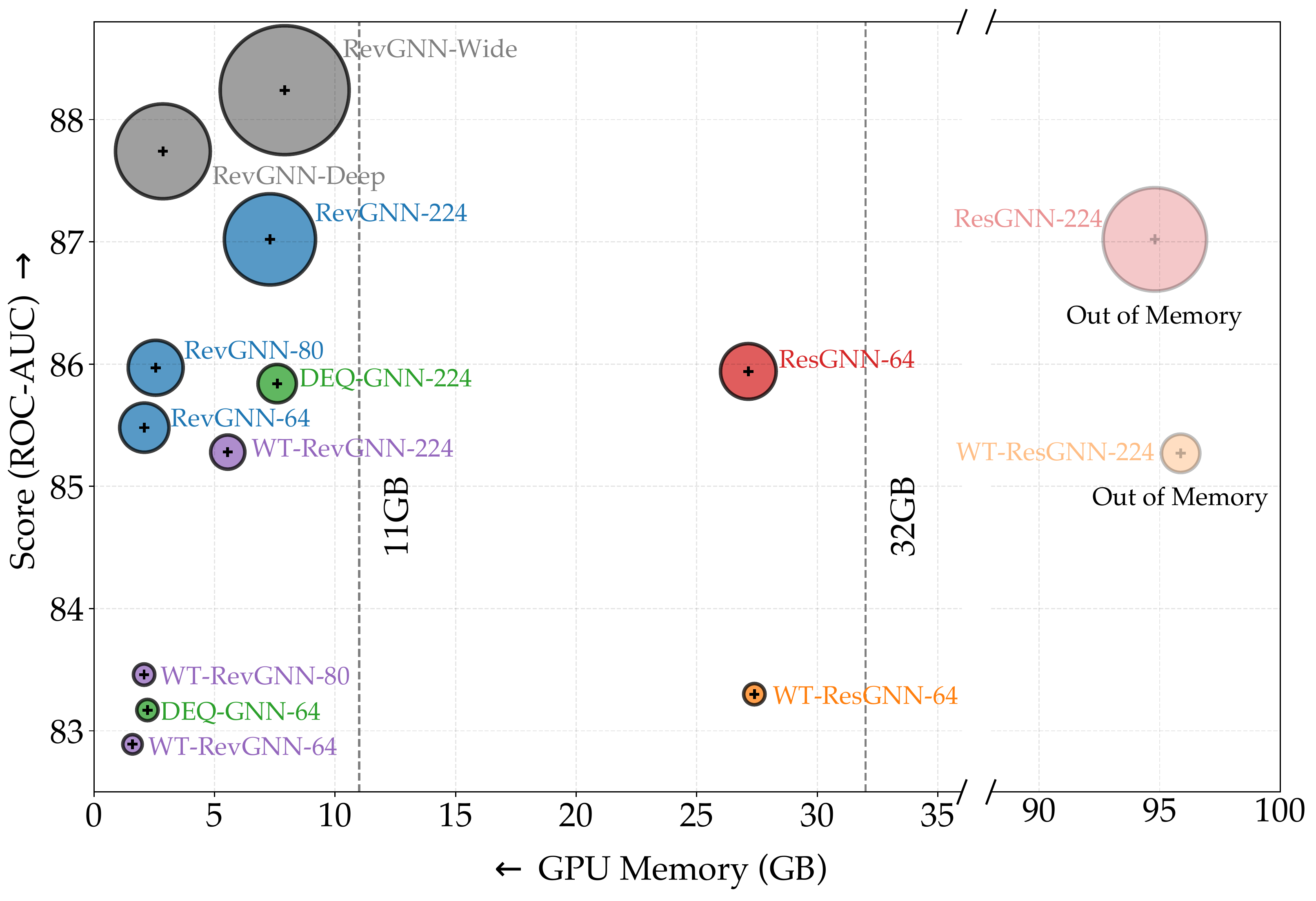}
    \caption{\textbf{ROC-AUC score \vs GPU memory consumption on the \emph{ogbn-proteins} dataset}. We find that deep reversible GNNs are very powerful and outperform existing models by a margin; our best models are RevGNN-Deep and RevGNN-Wide. We also compare reversible connections (RevGNN-x), weight-tying (WT-x), and equilibrium models (DEQ-x) for 112-layer deep GNNs (x denotes the number of channels per layer). Reversible models consistently achieve the same or better performance as the baseline using only a fraction of the memory. Weight-tied and equilibrium models offer a good performance to parameter efficiency trade-off. Datapoint size is proportional to $\sqrt{p}$, where $p$ is the number of parameters. 
    }
    \label{fig:memory_performance_obgn_proteins}
    \vspace{-16pt}
\end{figure}

Graphs are all around us. 
Whether we watch a show on Netflix, browse through friends' feeds on Facebook, buy something on Amazon, or look up a researcher on Google, chances are that our actions trigger queries on a large graph. Movie and book databases are often encoded as knowledge graphs for efficient recommendation systems. Social media services rely on social graphs. Shopping platforms leverage product co-purchasing networks to boost sales. Citation indices like Web of Science, Scopus, and Google Scholar construct large citation graphs. Even the Internet itself is in essence a vast graph with billions of nodes and edges. 
Graphs are also powerful tools for representing 3D data such as point clouds and meshes or biological data such as molecular structures and protein interactions.

One prominent and powerful approach to process such graphs are graph neural networks (GNNs). GNNs have achieved impressive performance on relatively small graph datasets \citep{yang2016revisiting,zitnik2017predicting,shchur2018pitfalls}.
Unfortunately, the most interesting and impactful real-world problems rely on very large graphs where limited GPU memory quickly becomes a bottleneck. In order to train GNN models on large graphs, the common practice is to reduce the number of model parameters. This is counterproductive, since processing larger graphs would likely benefit from more parameters.  
While there is evidence that over-parameterized models generalize better \citep{neyshabur2018towards,Belkin2019}, the relationship between performance and parameters is best illustrated in the context of language modelling. Recent progress in natural language processing (NLP) has been enabled by a massive increase in parameter counts: GPT (110M) \citep{radford2018improving}, BERT (340M) \citep{devlin2018bert}, GPT-3 (175B) \citep{brown2020language}, Gshard-M4 (600B) \citep{lepikhin2020gshard}, and DeepSpeed (1T) \citep{rasley2020deepspeed}. More parameters mean deeper or wider networks that consume more memory.

GNNs have shown a lot of promise on recent large-scale graph datasets such as Benchmarking GNNs \citep{dwivedi2020benchmarkgnns}, Open Graph Benchmark (OGB) \citep{hu2020open,hu2021ogblsc} and Microsoft Academic Graph (MAG) \citep{wang2020microsoft}. Recent works \citep{li2019deepgcns,li2020deepergcn,chen2020simple} have successfully trained deep models with a large number of parameters and achieved state-of-the-art performance. 
However, these models have large memory footprints and operate at the physical limits of current hardware. 
In order to apply deeper and wider GNNs with more parameters, we need either different hardware or more efficient architectures that consume less memory.

Existing works try to overcome the memory bottleneck by mini-batch training, \ie sampling a smaller set of nodes \citep{hamilton2017inductive,chen2018stochastic,chen2018fastgcn} or partitioning large graphs \citep{chiang2019cluster,graphsaint-iclr20} into smaller subgraphs and sampling from those. These approaches have proven successful, but they introduce further hyperparameters that need to be tuned. For instance, if the sampled size of nodes or subgraphs is too small, it may break important structures in the graph. While these methods are a step in the right direction, they do not scale well as the models become deeper or wider, since memory consumption is still dependent on the number of layers.
Another approach is efficient propagation via $K$-power adjacency matrices or graph diffusion matrices \citep{wu2019simplifying,klicpera_predict_2019,bojchevski2020scaling,liu2020towards,sign_icml_grl2020}. However, the propagation schemes of these methods are non-trainable, which may lead to sub-optimality.

Inspired by efficient architectures from computer vision and natural language processing \citep{gomez2017reversible,xie2017aggregated,bai2018trellis,bai2019deep}, here we investigate several methods to obtain more efficient GNNs that use less memory than conventional architectures while achieving state-of-the-art results (see \figLabel \ref{fig:memory_performance_obgn_proteins}). We explore grouped reversible graph connections in order to reduce the memory complexity with respect to the number of layers from $\mathcal{O}(L)$ to $\mathcal{O}(1)$. In other words, the memory consumption is independent of the depth. This allows us to train very deep, over-parameterized models with constant memory consumption. 

We also investigate parameter-efficient architectures. We study deep weight-tied GNNs that have the parameter cost of only a single layer. We also develop a deep graph equilibrium GNN, which is essentially a weight-tied network with infinite depth. We directly solve for the equilibrium point of this infinite-layer network using a root-finding method. We backpropagate through the equilibrium point using implicit differentiation. Hence, we do not need to store intermediate states and get an infinite-depth network at the memory and parameter cost of a single layer.

Our analysis of these methods shows that deep reversible architectures are the most powerful in terms of achieving state-of-the-art performance on benchmark datasets. This is due to their very large capacity at low memory cost and only slightly increased training time. 
Weight-tied models offer constant parameter size regardless of depth.
However, due to the smaller number of parameters, performance on large datasets suffers and has to be compensated by increasing the width. Finally, graph equilibrium models have the same memory efficiency as reversible models and the same parameter efficiency as weight-tied models. They perform similarly to weight-tied models, and the training time \vs performance tradeoff can be further adjusted by tuning the number of iterations in each optimization step.

Our methods can be applied to different GNN operators. In our experiments, we successfully apply them to GCN \citep{kipf2017semi}, GraphSAGE \citep{hamilton2017inductive}, GAT \citep{veli2018gat}, and DeepGCN \citep{li2019deepgcns}. We can also combine the proposed techniques with sampling-based methods to further reduce memory and boost performance on some datasets.
To the best of our knowledge, we are the first to train a GNN with more than $1000$ layers. Our model RevGNN-Deep, outperforms all state-of-the-art approaches on the \emph{ogbn-proteins} dataset \citep{hu2020open} with an ROC-AUC of 87.74\%, while only consuming 2.86 GB of GPU memory during training, one order of magnitude less than the current top performer. We can also trade our memory savings for larger width, pushing performance to new heights. Our RevGNN-Wide achieves an ROC-AUC of 88.24\% on the \emph{ogbn-proteins} dataset, ranking first on the leaderboard by a large margin. 

In summary, we investigate several techniques to increase the memory efficiency of GNNs and perform an extensive analysis. We significantly outperform current state-of-the-art methods on several datasets by employing reversible connections to train deeper and wider models. Further, we demonstrate the generality of these techniques by applying them to multiple GNN operators. We release our implementation, which supports PyTorch Geometric \citep{Fey/Lenssen/2019} and the Deep Graph Library 
\citep{wang2019dgl}.

\section{Related Work}
\label{sec:related}
GNNs were introduced by \citet{gori2005new} and \citet{scarselli2008graph} for learning a graph representation by finding stable states through fixed-point iterations. \citet{bruna2013spectral} generalized Convolutional Neural Networks (CNNs) to graphs using the Fourier basis by computing the eigen-decomposition of the graph Laplacian. 
Follow-up works \citep{ henaff2015deep,defferrard2016convolutional, kipf2017semi,levie2018cayleynets, li2018adaptive} propose different ways to improve and extend spectral GNNs. For instance, \citet{kipf2017semi} simplify spectral graph convolutions by limiting the filters to operate on 1-hop neighborhoods. Instead of defining GNNs from the spectral view, \citet{gilmer2017neural} introduce a general framework termed Message Passing Neural Networks (MPNNs) to unify GNNs. MPNNs define convolutions on the graph by propagating messages from spatial neighbors. Many subsequent works \citep{hamilton2017inductive, monti2017geometric, niepert2016learning, gao2018large, xu2018powerful, veli2018gat, wang2019dynamic} fall into this framework.

Most prior state-of-the-art (SOTA) works are limited to shallow depths due to the over-smoothing \citep{li2018deeper} and vanishing gradient problems \citep{li2019deepgcns,li2021deepgcns_pami} present in GNNs. Current works have explored different approaches to resolve these difficulties in training deep GNNs. The first line of research employs skip connections across layers. \citet{xu2018jknet} propose jump connections to adaptively select intermediate representations to the last layer. \citet{li2019deepgcns} adapt residual and dense connections \citep{he2016deep, huang2017densely} and dilated convolutions \citep{YuKoltun2016} from CNNs to GNNs; they successfully train GNNs with up to 56 layers on 3D point clouds and achieve SOTA results. \citet{li2020deepergcn} train a 112-layer residual GNN and achieve SOTA performance on the large-scale \emph{ogbn-proteins}  dataset \citep{hu2020open}. 
Other works also demonstrate that residual connections aid in training deeper GNNs \citep{gong2020geometrically,chen2020simple,xu2021optimization}.
We push further along the path of deep residual GNNs and train a SOTA model with more than 1000 layers by addressing the memory bottleneck of current approaches.

Researchers have also studied different normalization and regularization techniques such as DropEdge \citep{rong2020dropedge}, DropConnect \citep{hasanzadeh2020bayesian}, PairNorm \citep{zhao2019pairnorm}, weight normalization \citep{oono2019graph}, differentiable group normalization \citep{zhou2020towards}, and GraphNorm \citep{cai2020graphnorm}. Another line of work \citep{wu2019simplifying,klicpera_predict_2019,bojchevski2020scaling,liu2020towards,sign_icml_grl2020} proposes an efficient propagation scheme to avoid stacking layers by incorporating multi-hop information into a single GNN layer via $K$-th power adjacency matrices or personalized PageRank diffusion matrices. However, the propagation scheme of these methods is non-trainable, thus making it difficult to learn hierarchical features and  limiting capacity.

Many applications yield huge graphs with millions of nodes and edges. Memory limitations preclude full-batch training in such settings. Specifically, to train an $L$-layer GNN with $D$ hidden channels on a graph $\mathcal{G}$ with $N$ nodes and $M$ edges, the memory complexity of storing activations is $\mathcal{O}(LND)$. GraphSAGE \citep{hamilton2017inductive} incorporates a recursive node-wise sampling scheme to improve the scalability of GNNs. Instead of training on the whole graph, GraphSAGE uniformly samples a fixed number of neighbors for a batch of nodes. However, the recursive neighborhood leads to exponential memory complexity as the number of GNN layers increases. If $B$ is the batch size of nodes and $R$ is the number of sampled neighbors per node, the memory complexity of GraphSAGE is $\mathcal{O}(R^{L}BD)$. VR-GCN \citep{chen2018stochastic} enhances GraphSAGE by reducing the variance of mini-batch training to increase the convergence rate. However, to reduce the estimation variance, historical activations need to be stored in memory ($\mathcal{O}(LND)$). 

To avoid recursive neighborhood expansion, FastGCN \citep{chen2018fastgcn} performs layer-wise sampling to subsample nodes for each layer independently with a degree-based distribution. While this further reduces the memory complexity to $\mathcal{O}(LRBD)$, it leads to sparse connections.
To better maintain the graph structure, Cluster-GCN \citep{chiang2019cluster} and GraphSAINT \citep{graphsaint-iclr20} propose subgraph-wise sampling methods for mini-batch training with memory complexity $\mathcal{O}(LBD)$. However, even with smart sampling methods, GNNs still do not scale as the number of layers increases (\ie the $L$ dimension). We tackle this issue and study orthogonal approaches that eliminate the $L$ dimension from the memory complexity. Our techniques enable unlimited network depth at no memory cost and can be combined with existing sampling methods.

\section{Building Deep GNNs}
\subsection{Preliminaries}
A graph $\mathcal{G}$ is represented by a tuple $\mathcal{G} = \tuple{\mathcal{V}, \mathcal{E}}$, where $\mathcal{V} = \set{v_1, v_2, ..., v_N}$ is an unordered set of vertices and $\mathcal{E} \subseteq \mathcal{V} \times \mathcal{V}$ is a set of edges. Let $N$ and $M$ denote the number of vertices and edges, respectively. For convenience, a graph can be equivalently defined as an adjacency matrix ${A \in \mathcal{A} \subset \mathbb{R}^{N\times N}}$, where $a_{i,j}$ denotes the link relation between vertex $v_{i}$ and $v_{j}$. In some scenarios, vertices and edges are associated with a vertex feature matrix $X \in \mathcal{X} \subset \Re^{N\times D}$ and an edge feature matrix $U \in \mathcal{U} \subset \mathbb{R}^{M\times F}$, respectively.
We use GNN operators that map the vertex feature matrix $X$, the adjacency matrix $A$, and the edge feature matrix $U$ (optional) to a transformed vertex feature matrix $X^{\prime}$:
\begin{align}
f_{w}: \mathcal{X} \times \mathcal{A} \times \mathcal{U} \mapsto \mathcal{X^{\prime}},
\end{align}
where $f_{w}(X, A, U)$ is parameterized by learnable parameters $w$. For simplicity, we assume that the transformed vertex feature matrix $X^{\prime}$ has the same dimension as the input vertex feature matrix $X$. We also assume that the adjacency matrix $A$ is the same for all GNN layers. When the edge feature matrix $U$ is present, it is fed into each layer with its initial values $U^{(0)}$.

\subsection{Over-parameterized GNNs}
\mysection{Deep GNNs}  Recent works \citep{li2019deepgcns, li2020deepergcn} show how adding residual connections \citep{he2016deep} to vertex features ($X^{\prime} = f_{w}(X, A, U) + X$) enables training of deep GNNs that achieve promising results on graph datasets. 
However, the memory complexity of the activations remains $\mathcal{O}(LND)$, where $L$ is the number of GNN layers, $N$ is the number of vertices, and $D$ is the size of vertex features. Hence, the memory consumption of deep GNNs scales linearly with the number of layers. Since the memory footprint of the network parameters is usually negligible, we focus on memory consumption induced by the activations.

\mysection{Grouped Reversible GNNs} \label{grgnn}
Inspired by reversible networks \citep{gomez2017reversible,liu2019gnf, kitaev2019reformer}
and grouped convolutions \citep{krizhevsky2012imagenet, xie2017aggregated}, we generalize reversible residual connections to grouped reversible residual connections for GNNs. 
Specifically, the input vertex feature matrix $X$ is uniformly partitioned into $C$ groups
$\tuple{X_{1}, X_{2}, ..., X_{C}}$ across the channel dimension, where $X_{i} \in \Re^{N\times \frac{D}{C}}$.
A grouped reversible GNN block operates on a group of inputs and produces a group of outputs: $\tuple{X_{1}, X_{2}, ..., X_{C}} \mapsto \tuple{X_{1}^{\prime}, X_{2}^{\prime}, ..., X_{C}^{\prime}}$. The forward pass is defined as follows:
\begin{align}
X_{0}^{\prime} &= \sum_{i=2}^{C} X_{i} \\
X_{i}^{\prime} &= f_{w_i}(X_{i-1}^{\prime}, A, U) + X_{i},~ i \in \{1,\cdots,C\},
\end{align}
where $X_{0}^{\prime}$ is designed for exchanging information across groups. Unlike conventional GNNs, grouped reversible GNNs only need to save the output vertex features of the last GNN block in GPU memory for backpropagation. Therefore, the memory complexity of activations becomes $\mathcal{O}(ND)$, which is independent of the depth of the network. Note that the adjacency matrix $A$ and the edge feature matrix $U$ are not updated during message passing. During the backward pass, only the input vertex features are reconstructed, on the fly, from the output vertex features $\tuple{X_{1}^{\prime}, X_{2}^{\prime}, ..., X_{C}^{\prime}}$ for backpropagation:
\begin{align}
X_{i} &= X_{i}^{\prime} - f_{w_{i}}(X_{i-1}^{\prime}, A, U),~ i \in \{2,\cdots,C\} \\
X_{0}^{\prime} &= \sum_{i=2}^{C} X_{i} \\
X_{1} &= X_{1}^{\prime} - f_{w_{1}}(X_{0}^{\prime}, A, U).
\end{align}
In practice, $X_{i}$ for $i \in\{2,\cdots,C\}$ can be computed in parallel. To reconstruct $X_{1}$, $X_{0}^{\prime}$ needs to be computed in advance. After reconstructing the original input vertex features, gradients can be derived through backpropagation. Owing to the group processing, the number of parameters reduces as the group size increases. Note that in the special case where the group size $C=2$, we obtain a similar form to the reversible residual connections proposed for CNNs \citep{gomez2017reversible}.
The definition above is independent of the choice of $f_{w}$. However, we find that normalization layers and dropout layers are essential for training deep GNNs. To avoid extra memory usage, normalization layers and dropout layers are embedded into the reversible GNN block. The GNN block $f_{w_{i}}$ is designed similar to the pre-activation residual GNN block proposed by \citet{li2020deepergcn}:
\begin{align}
\widehat{X}_{i} &= \text{Dropout}(\text{ReLU}(\text{Norm}(X_{i-1}^{\prime}))) \\
\widetilde{X}_{i} &= \text{GraphConv}(\widehat{X}_{i}, A, U).
\end{align}
The stochasticity of vanilla dropout layers causes reconstruction errors in the reverse pass. A naive solution would be to store the dropout pattern for all layers. However, the dropout patterns have the same dimension as the activations, which would lead to $\mathcal{O}(LND)$ memory consumption.
As an alternative, we adopt a modified dropout layer in which the dropout pattern is shared across layers. Therefore, we only need to store one dropout pattern in every SGD iteration; its memory complexity is independent of the depth: $\mathcal{O}(ND)$. During the reverse pass, the saved dropout pattern is reactivated to reconstruct the input vertex features.

\subsection{Parameter-Efficient GNNs}
\mysection{Weight-tied GNNs}
Weight-tying is a powerful tool for improving the parameter efficiency of language models \citep{press2017using,inan2016tying,bai2018trellis}. We take these works as inspiration to study how weight-tying can improve the parameter efficiency of GNNs. Instead of having different weights for different GNN blocks, the weight-tied GNN shares weights across layers. The GNN model proposed by \citet{scarselli2008graph} can be considered the first weight-tied GNN, in which a learned transition function is used to find stable node states by Banach's fixed point theorem \citep{khamsi2001introduction}. Here we experiment with weight-tied residual GNNs and weight-tied reversible GNNs. For weight-tied residual GNNs, we define:
\begin{align}
f_{w}^{(1)} := f_{w}^{(2)} \ldots := f_{w}^{(L)},
\end{align}
where $L$ is the number of layers. For weight-tied reversible GNNs, weights are shared in a group-wise manner:
\begin{align}
f_{w_{i}}^{(1)} := f_{w_{i}}^{(2)} \ldots := f_{w_{i}}^{(L)},~ i \in \{1,\cdots,C\} 
\end{align}
Note that we use the same pre-activation GNN block described in \secLabel \ref{grgnn} instead of a contraction map used by \citet{scarselli2008graph}. Both weight-tied GNNs have explicit layers and are trained with backpropagation. But the weight-tied reversible GNN reconstructs input vertex features on the fly during backpropagration. Therefore, the memory complexities of the weight-tied residual GNN and weight-tied reversible GNN are $\mathcal{O}(LND)$ and $\mathcal{O}(ND)$, respectively.

\mysection{Deep Equilibrium GNNs}
An alternative way to train weight-tied GNNs with $\mathcal{O}(ND)$ memory consumption is to use implicit differentiation %\citep{scarselli2008graph, bai2019deep}, 
\citep{scarselli2008graph, bai2019deep}, % gu2020implicit
which assumes that the network can reach an equilibrium state. We construct a GNN model that is assumed to converge to a fixed point $Z^{*}$ for any given input:
\begin{align}
Z^{*} &= f_{w}^{\text{DEQ}}(Z^{*}, X, A, U),
\end{align}
where the state $Z$ represents the transformed node features. To construct a stable or contractive GNN block, we mimic the design of MDEQ \citep{bai2020multiscale}. We build a GNN block as follows:
\begin{align}
& Z^{\prime} = \text{GraphConv}(Z_{\text{in}}, A, U) \\
& Z^{\mydprime} = \text{Norm}(Z^{\prime} + X) \\
& Z^{\mytrprime} = \text{GraphConv}(\text{Dropout}(\text{ReLU}(Z^{\mydprime})), A, U) \\
& Z_{\text{o}} = \text{Norm}(\text{ReLU}(Z^{\mytrprime} + Z^{\prime})),
\end{align}
where $Z_{\text{in}}$ is the input node state, $Z_{\text{o}}$ is the output node state, $X$ serves as the injected input, and $Z^{\prime}$ forms an internal residual signal to the output $Z_{\text{o}}$. In practice, $X$ represents the initial node features and $Z_{\text{in}}$ is initialized to zeros for the first iteration. Similar to DEQ \citep{bai2019deep}, the forward pass of DEQ-GNN is implemented with a root-finding algorithm (\eg Broyden’s method) and the gradients are obtained by implicitly differentiating through the equilibrium node state for the backward pass.

\subsection{Analysis of Different Deep GNN Architectures} \label{sec:analysis_deep}
In the following, we compare the different techniques for building deep GNNs with respect to their performance, memory efficiency and parameter efficiency on the \emph{ogbn-proteins} dataset from the \emph{Open Graph Benchmark} (OGB) \citep{hu2020open}. We use the same GNN operator \citep{li2020deepergcn}, hyper-parameters (\eg learning rate, dropout rate, training epoch, \etc), and optimizers to make the comparison as fair as possible.
We use mini-batch training with random partitioning where graphs are split into 10 parts during training and 5 parts during testing \citep{li2020deepergcn}. One subgraph is sampled to form a mini-batch at each step. 
The reported GPU memory corresponds to the peak GPU memory usage during the first training epoch.

\begin{figure}[!t]
    \centering
    \includegraphics[trim=0cm 0cm 0cm 0cm, width=\columnwidth]{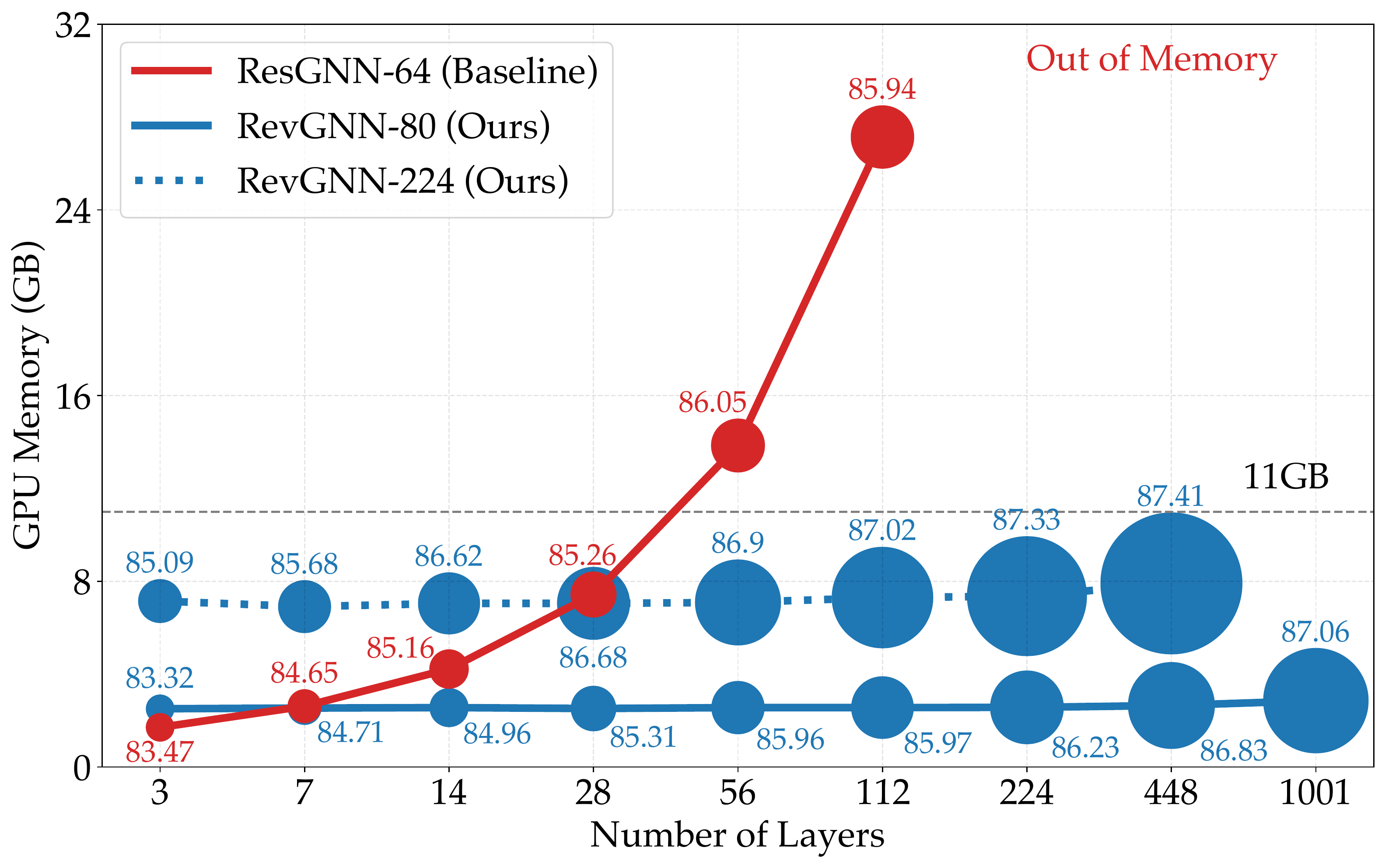}
    \caption{\textbf{GPU memory consumption \vs number of layers for ResGNN \citep{li2020deepergcn} and RevGNN, our adaptation with reversible connections.} RevGNN uses constant memory for deeper networks with more parameters and better performance. We use width $64$ and $80$ for ResGNN and RevGNN, respectively, to ensure a similar number of parameters per network depth. Datapoints are annotated with the ROC-AUC score of the model and their size is proportional to $\sqrt{p}$, where $p$ is the number of parameters.}
    \label{fig:rev_memory_layers}
    %\vspace{-8pt}
\end{figure}

\mysection{Baseline GNN}
We use a recent pre-activation residual GNN \citep{li2020deepergcn}, which achieves state-of-the-art performance across several OGB datasets, as our baseline. For simplicity, we use the \emph{max} aggregation function for all the ablated models on the \emph{ogbn-proteins} dataset. We refer to the baseline model as ResGNN. As shown in \figLabel \ref{fig:rev_memory_layers}, the ResGNN with 112 layers and a channel size of 64 achieves $85.94\%$ ROC-AUC on \emph{ogbn-proteins}. However, the memory consumption of ResGNN-64 increases linearly with the number of layers. ResGNN-64 runs out of memory beyond 112 layers, making it impossible to investigate deeper models with current commodity hardware.

\mysection{Reversible GNN}
Our reversible GNN enables training of very deep networks with constant memory footprint, as illustrated in \figLabel \ref{fig:rev_memory_layers} (RevGNN). We use a group size of $2$; since grouping reduces the number of parameters, RevGNN-80 has roughly the same number of parameters as ResGNN-64 for the same number of layers. While the baseline model ResGNN-64 cannot go beyond 112 layers due to memory limitations, RevGNN-80 can go to more than 1000 layers without additional memory cost and achieves much better accuracy (87.06\% ROC-AUC). 
We can invest the saved GPU memory to increase the network width and train a higher-capacity model. This model is not only deep (448 layers) but also wide (224 channels) and further improves performance (87.41\% ROC-AUC).

\begin{figure}[h]
    \centering
    \includegraphics[trim=0cm 1cm 0cm 0cm, width=\columnwidth]{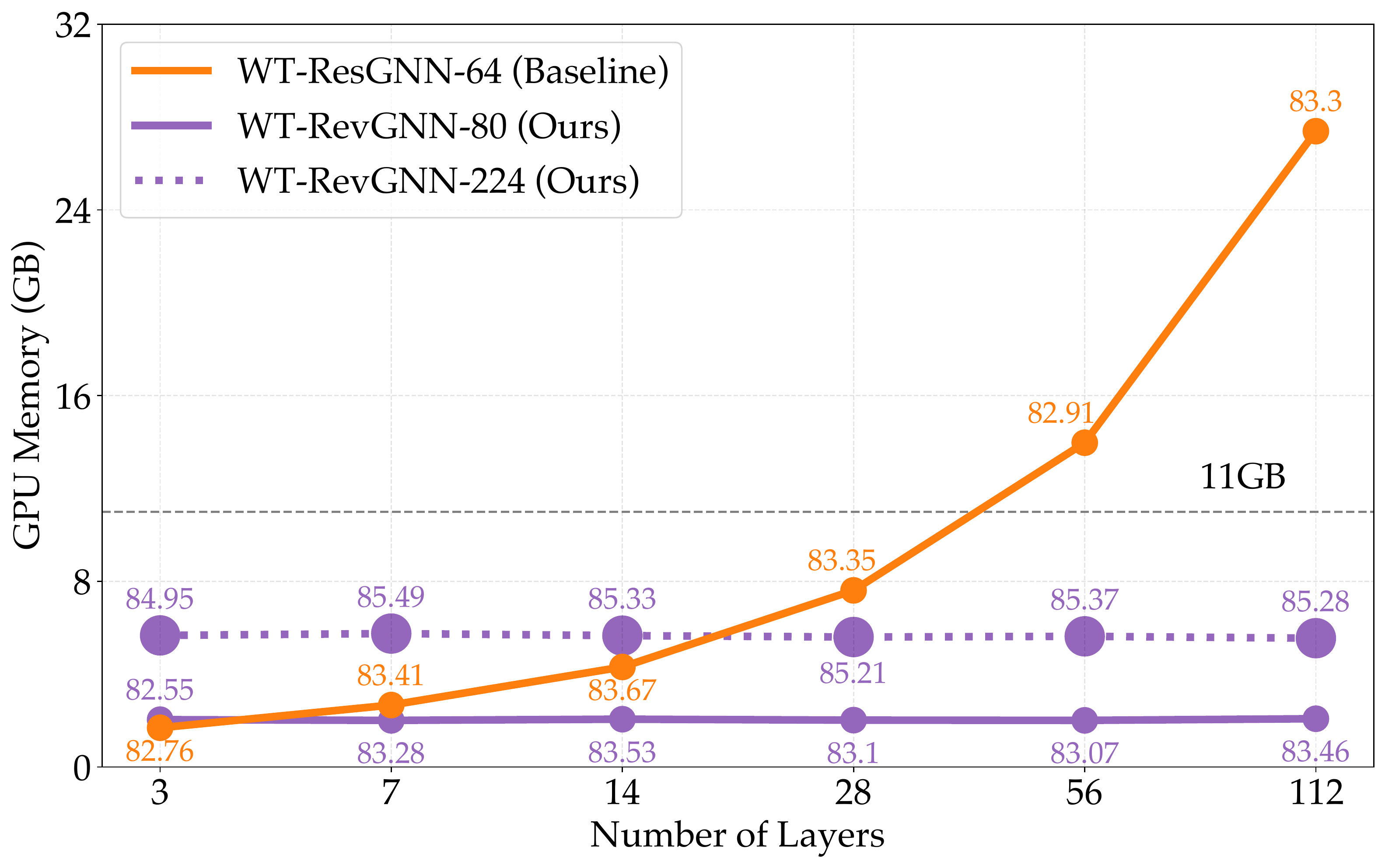}
    \caption{\textbf{GPU memory consumption \vs number of layers for WT-ResGNN and WT-RevGNN.} For a given filter size, the number of parameters is constant regardless of network depth. Our reversible architecture has a constant memory footprint regardless of network depth, while yielding similar accuracy. Datapoints are annotated with the ROC-AUC score of the model and their size is proportional to $\sqrt{p}$, where $p$ is the number of parameters.}
    \label{fig:memory_layers_wt}
    \vspace{4pt}
\end{figure}

\mysection{Weight-tied GNN}
We compare the weight-tied ResGNN (WT-ResGNN) and its reversible counterpart RevGNN (WT-RevGNN) in \figLabel \ref{fig:memory_layers_wt}. Both models have approximately the same number of parameters and achieve a similar accuracy while the reversible model uses significantly less GPU memory. Since the parameters are shared across GNN layers, the number of parameters stays constant as the number of layers increases. For example, with 112 layers, WT-RevGNN-224 has only 337k parameters, while RevGNN-224 has 17.1M parameters. However, training time and GPU memory consumption are similar while the performance of the weight-tied model is worse (85.28\% \vs 87.41\%) due to diminishing returns after more than 7 layers. Similar to the results in \figLabel \ref{fig:rev_memory_layers}, this is evidence of a clear correlation between the number of parameters and performance.  

\begin{figure}[h]
    \centering
    \includegraphics[trim=0cm 1cm 0cm 0cm, width=\columnwidth]{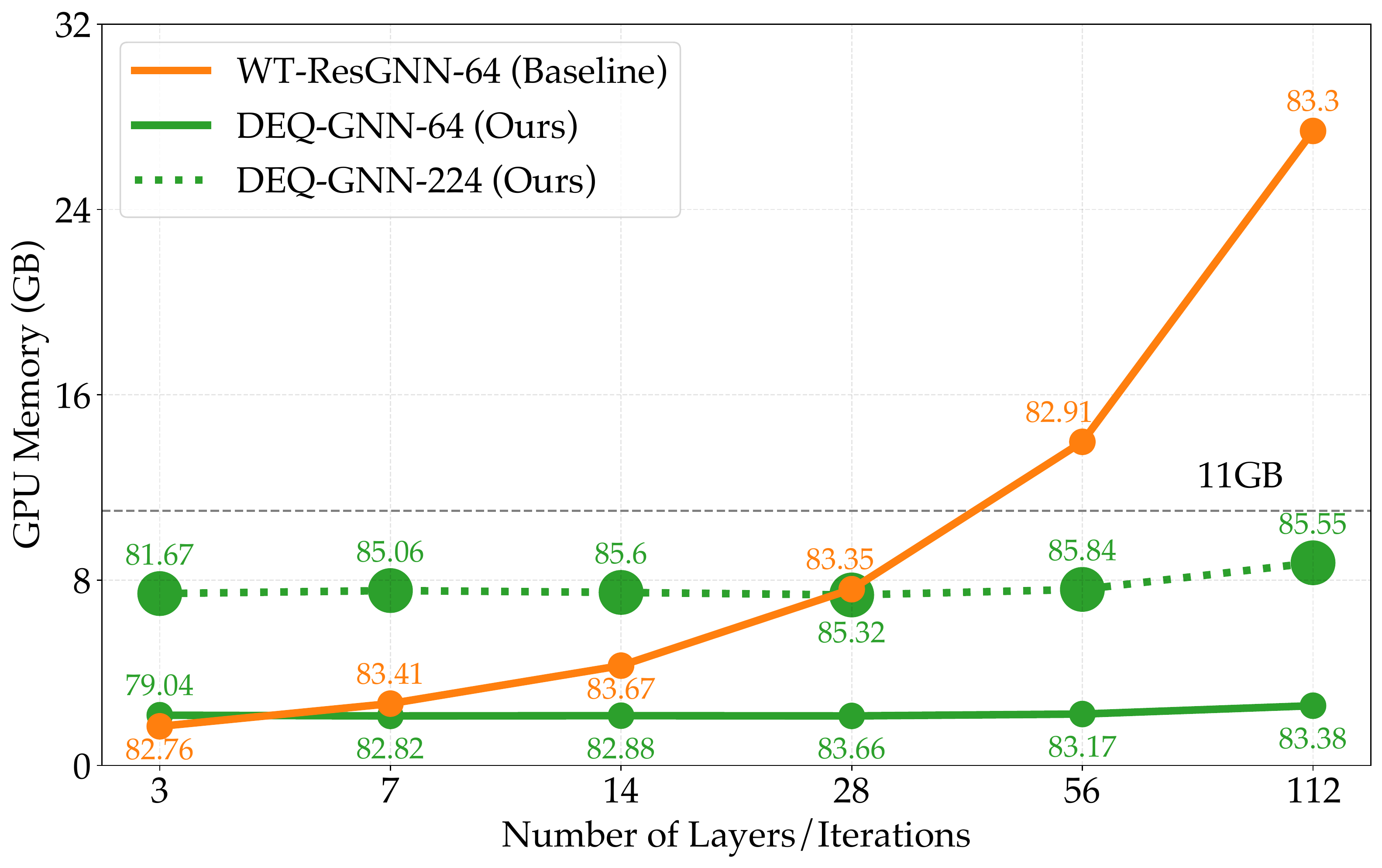}
    \caption{\textbf{GPU memory \vs number of layers/iterations for ResGNN and DEQ-GNN.} Datapoints are annotated with the ROC-AUC score of the model and their size is proportional to $\sqrt{p}$, where $p$ is the number of parameters.}
    \label{fig:memory_iterations_deq}
    \vspace{4pt}
\end{figure}

\mysection{Equilibrium GNN}
Equilibrium networks implicitly model a weight-tied network with infinite depth. As a result, they only have the parameter and memory footprint of a single layer, but the expressiveness of a very deep network. The initial node features $X$ are used as input injection and the initial node states $Z$ are set to zero. We implement our DEQ-GNN based on the original DEQ implementation for CNNs \citep{bai2019deep}. Broyden’s root-finding method is used \citep{broyden1965class} in both the forward pass and the backward pass to find the equilibrium node states and approximate the inverse Jacobian. The Broyden iterations terminate when the norm of the objective is smaller than a tolerance $\epsilon$ or a maximum iteration threshold is reached. $\epsilon$ is set to $10^{-6} \cdot \sqrt{BD}$ and $2 \times 10^{-10} \cdot \sqrt{BD}$ for the forward pass and the backward pass respectively, where $B$ is the number of nodes in the sampled subgraph and $D$ is the channel size. The iteration thresholds in the forward pass and the backward pass are set to the same value. We examine different iteration thresholds for DEQ-GNN with a channel size of 64 (DEQ-GNN-64) and a channel size of 224 (DEQ-GNN-224) in \figLabel \ref{fig:memory_iterations_deq}. It can be seen that the iteration threshold is crucial for good performance, since it affects the convergence to the equilibrium. We find that DEQ-GNN-64 performs similarly to WT-RevGNN-80 with nearly the same number of parameters and memory consumption.
The wider DEQ-GNN-224 model reaches $85.84\%$ ROC-AUC, which is comparable to the 112-layer ResGNN-64, with only around $28\%$ memory footprint and $23\%$ parameters (Mem: 7.60GB \vs 27.1GB, Params: 537k \vs 2.37M). DEQ-GNN-224 slightly outperforms WT-RevGNN-224 with only about 60\% of the training time.

\mysection{Discussion}
Reversible networks emerge as the most promising approach for building deep GNNs that achieve state-of-the-art performance. They consume much less memory than the baseline while achieving comparable performance with the same number of parameters (see \figLabel \ref{fig:memory_performance_obgn_proteins}). However, in contrast to the baseline method, we are able to train very deep networks with more parameters and much better performance without running out of memory. While it is possible to go to arbitrary depths with constant memory consumption, the training time increases. In order to reduce the number of parameters and inference time, it is possible to increase the group size. However, we find that group sizes larger than 2 do not lead to a performance increase in practice and may even degrade model performance. We conjecture that this is due to the increased receptive field of early layers and the smaller number of parameters per memory. We provide an ablation study in the appendix.

The weight-tied network limits the number of parameters to a single layer regardless of the effective depth. We find that going deeper with tied weights boosts performance, but returns eventually diminish (see \figLabel \ref{fig:memory_layers_wt}). While parameters stay constant, going deeper still increases memory consumption, unless the reversible GNN block is used. 
An extension of the weight-tied network is the graph equilibrium model. It represents an infinite-depth weight-tied network and uses fixed-point iterations to solve for the optimal weights. This allows for much wider channel size with the same amount of memory. DEQ-GNNs are faster to train, but have more hyper-parameters to tune. Please refer to the appendix for a training time ablation. We find that pretraining is not necessary, but can further improve the results. The results reported in this paper are without pretraining for fair comparison. While weight-tied GNNs and equilibrium GNNs are not able to achieve state-of-the-art performance, they are very parameter-efficient, which may be relevant for applications on embedded devices where model size matters. 

\section{Over-parameterized Deep GNNs}
We conduct experiments on several datasets from the \emph{Open Graph Benchmark} (OGB) \citep{hu2020open}. We first show state-of-the art results on several datasets with the proposed deep reversible GNN. We then apply the reversible GNN framework to different GNN operators on the \emph{ogbn-arxiv} dataset. To show how mini-batch sampling can further aid the training of deep GNNs, we compare full-batch and mini-batch training of reversible GNNs on the \emph{ogbn-products} dataset.
The data splits and evaluation metrics on all datasets follow the OGB evaluation protocol. Mean and standard deviation are obtained across 10 trials. All the ablated models use the same hyper-parameters (\eg learning rate, dropout rate, training epoch, \etc) and optimizers as the baseline models. The implementation of all the reversible models is based on PyTorch \citep{pytorch2019} and supports both PyTorch Geometric (PyG) \citep{Fey/Lenssen/2019} and Deep Graph Libray (DGL)~\citep{wang2019dgl} frameworks.

\subsection{State-of-the-art Results}
We briefly describe two variants of our RevGNN with reversible connections, which reach new SOTA results on the \emph{ogbn-proteins} dataset of the OGB leaderboard \citep{hu2020open} (\tblLabel \ref{table:ogbn_proteins_sota}) at the time of submission. RevGNN-Deep has 1001 layers and a channel size of 80. It outperforms the previous SOTA method UniMP+CEF by 0.83\% ROC-AUC, while using only 10.5\% of the GPU memory for training. RevGNN-Wide uses 448 GNN layers and 224 hidden channels and significantly outperforms UniMP+CEF by 1.33\% ROC-AUC with about 29\% of the GPU memory. RevGNN-Deep uses the same training setting as mentioned in \secLabel \ref{sec:analysis_deep}. The RevGNN-Wide uses a larger dropout rate of $0.2$ to prevent overfitting. To boost the performance, the results of RevGNN-Deep and RevGNN-Wide are obtained using multi-view inferences with 10 views on larger subgraphs with a partition size of 3. We perform the inferences on a NVIDIA RTX A6000 (48GB). Please refer to the appendix for the details of multi-view inference.
Larger and deeper models incur a cost in terms of training and inference time. RevGNN-Deep and RevGNN-Wide take 13.5 days and 17.1 days, respectively, to train for 2000 epochs on a single NVIDIA V100. Nonetheless, it is affordable for accuracy-critical applications in scientific research such as predicting protein structures \citep{senior2020improved}. We demonstrate that it is possible to train huge over-parameterized GNN models on a single GPU. The RevGNN-Wide model has 68.47 million parameters, which is about half the size of the GPT language model \citep{radford2018improving}. We believe that this is an important step forward in developing over-parameterized GNNs for graphs.

Our RevGNNs also achieve new SOTA results on the \emph{ogbn-arxiv} dataset (see \tblLabel \ref{table:ogbn_arxiv_sota}). RevGCN-Deep uses 28 GCN \citep{kipf2017semi} layers with 128 channels each and achieves an accuracy of 73.01\%, while  only using 1.84GB of GPU memory.
RevGAT-Wide uses 5 attention layers with 3 heads and 356 channels for each head and outperforms the current top performer UniMP\_v2 \citep{shi2020masked} ($74.05\%$ \vs $73.97\%$) while using about a third of the memory (8.49GB \vs 25.0GB). RevGAT-SelfKD uses self-knowledge distillation \citep{Zhang2019be} with 5 attention layers, 3 heads, and 256 channels each. The teacher models achieve an accuracy of $74.02\%$. After training with distillation, the student models set a new SOTA with $74.26\%$ test accuracy. Please refer to the appendix for more details.

\begin{table}[t]
\vspace{-8pt}
\centering
\setlength{\tabcolsep}{1pt}
\caption{\textbf{Results on the \emph{ogbn-proteins} dataset compared to SOTA.} RevGNN-Deep has 1001 layers with 80 channels each. It achieves SOTA performance with minimal GPU memory for training. RevGNN-Wide has 448 layers with 224 channels each. It achieves the best accuracy while consuming a moderate amount of GPU memory.}
\vspace{2pt}
\begin{tabular}{cccccc}
\toprule
  \label{table:ogbn_proteins_sota}
  \centering
Model & ROC-AUC $\uparrow$
& Mem $\downarrow$
& Params \\
\midrule
GCN (\citeauthor{kipf2017semi}) & 72.51 \textcolor{gray}{\small{± 0.35}} & 4.68 
& 96.9k %96,880 
\\
GraphSAGE (\citeauthor{hamilton2017inductive}) & 77.68 \textcolor{gray}{\small{± 0.20}} & 3.12 
& 193k %193,136 
\\
DeeperGCN (\citeauthor{li2020deepergcn}) & 86.16 \textcolor{gray}{\small{± 0.16}} & 27.1
& 2.37M %2,374,568 
\\
UniMP (\citeauthor{shi2020masked}) & 86.42 \textcolor{gray}{\small{± 0.08}} & 27.2
& 1.91M %1,909,104 
\\
GAT (\citeauthor{veli2018gat}) & 86.82 \textcolor{gray}{\small{± 0.21}} & 6.74
& 2.48M %2,475,232 
\\
UniMP+CEF (\citeauthor{shi2020masked}) & 86.91 \textcolor{gray}{\small{± 0.18}} 
& 27.2
& 1.96M %1,959,984 
\\
\midrule
Ours (RevGNN-Deep)& 87.74 \textcolor{gray}{\small{± 0.13}} & \textbf{2.86} & 20.03M
\\
Ours (RevGNN-Wide) & \textbf{88.24} \textcolor{gray}{\small{± 0.15}} & 7.91 & 68.47M 
\\
\bottomrule
\end{tabular}
\end{table}

\begin{table}[!htb]
\vspace{-8pt}
\centering
\setlength{\tabcolsep}{1pt}
\caption{\textbf{Results on the \emph{ogbn-arxiv} dataset compared to SOTA.} RevGCN-Deep has 28 layers with 128 channels each. It achieves SOTA performance with minimal GPU memory. RevGAT-Wide has 5 layers with 1068 channels each. RevGAT-SelfKD denotes the student models with 5 layers and 768 channels. It achieves the best accuracy while consuming a moderate amount of GPU memory.}
\vspace{2pt}
\begin{tabular}{cccccc}
\toprule
  \label{table:ogbn_arxiv_sota}
  \centering
Model & ACC $\uparrow$
& Mem $\downarrow$
& Params \\
\midrule
GraphSAGE (\citeauthor{hamilton2017inductive}) & 71.49 \textcolor{gray}{\small{± 0.27}}
& 1.99
& 219k %218,664 
\\
GCN (\citeauthor{kipf2017semi}) & 71.74 \textcolor{gray}{\small{± 0.29}}
& 1.90 
& 143k %142,888 
\\
DAGNN (\citeauthor{liu2020towards}) & 72.09 \textcolor{gray}{\small{± 0.25}}
& 2.40 
& 43.9k %43,857
\\
DeeperGCN (\citeauthor{li2020deepergcn}) & 72.32 \textcolor{gray}{\small{± 0.27}}
& 21.6 
& 491k %491,176 
\\
GCNII (\citeauthor{chen2020simple}) & 72.74 \textcolor{gray}{\small{± 0.16}}
& 17.0
& 2.15M %2,148,648
\\
GAT (\citeauthor{veli2018gat}) & 73.91 \textcolor{gray}{\small{± 0.12}}
& 5.52
& 1.44M %1,441,580 
\\
UniMP\_v2 (\citeauthor{shi2020masked}) & 73.97 \textcolor{gray}{\small{± 0.15}}
& 25.0
& 687k %687,377 
\\
\midrule
% Ours (RevSAGE-Deep) & 72.73 ± 0.10
% & \textbf{1.57}
% & 953k %953,140 
Ours (RevGCN-Deep) & 73.01 \textcolor{gray}{\small{± 0.31}}
& \textbf{1.84}
& 262k %953,140 
\\
Ours (RevGAT-Wide) & 74.05 \textcolor{gray}{\small{± 0.11}} & 8.49  & 3.88M %3,879,056 
\\
Ours (RevGAT-SelfKD) & \textbf{74.26} \textcolor{gray}{\small{± 0.17}} & 6.60  & 2.10M
\\
\bottomrule
\end{tabular}
\end{table}

\begin{table}[t]
\vspace{-8pt}
\centering
\setlength{\tabcolsep}{5pt}
\caption{\textbf{Results with different GNN operators on the \emph{ogbn-arxiv}.} All GAT models use label propagation. \#L and \#Ch denote the number of layers and channels respectively. \textit{Baselines} are in italic.}
\vspace{2pt}
\begin{tabular}{cccccc}
\toprule
  \label{table:ogbn_arxiv_ablation}
  \centering
Model & \#L & \#Ch & ACC $\uparrow$ & Mem $\downarrow$ & Params \\
\midrule
\textit{ResGCN} & 28 & 128 & 72.46 \textcolor{gray}{\small{± 0.29}} & 11.15
& 491k %491,176 
\\
RevGCN & 28 & 128 & 73.01 \textcolor{gray}{\small{± 0.31}} & \textbf{1.84}
& 262k %262,056
\\
RevGCN & 28 & 180 & \textbf{73.22} \textcolor{gray}{\small{± 0.19}} & 2.73
& 500k %499,540 
\\
\midrule
\textit{ResSAGE} & 28 & 128 & 72.46 \textcolor{gray}{\small{± 0.29}} & 8.93
& 950k %949,928 
\\
RevSAGE & 28 & 128 & 72.69 \textcolor{gray}{\small{± 0.23}} & \textbf{1.17}
& 491k %491,432 
\\
RevSAGE & 28 & 180 & \textbf{72.73} \textcolor{gray}{\small{± 0.10}} & 1.57 
& 953k %953,140  
\\
\midrule
\textit{ResGEN} & 28 & 128 & 72.32 \textcolor{gray}{\small{± 0.27}} & 21.63 
& 491k %491,176  
\\
RevGEN & 28 & 128 & 72.34 \textcolor{gray}{\small{± 0.18}} & \textbf{4.08} 
& 262k %262,056  
\\
RevGEN & 28 & 180 & \textbf{72.93} \textcolor{gray}{\small{± 0.10}} & 5.67
& 500k %499,540 
\\
\midrule
\textit{ResGAT} & 5 & 768 & 73.76 \textcolor{gray}{\small{± 0.13}} & 9.96
& 3.87M %3,867,728
\\
RevGAT & 5 & 768 & 74.02 \textcolor{gray}{\small{± 0.18}} & \textbf{6.30}
& 2.10M %2,098,256
\\
RevGAT & 5 & 1068 & \textbf{74.05} \textcolor{gray}{\small{± 0.11}} & 8.49
& 3.88M %3,879,056
\\
\bottomrule
\end{tabular}
\end{table}

\subsection{Application to Different GNN Operators}
The proposed techniques are generic and can in principle be applied to any SOTA GNN to further boost the performance with deeper and wider architectures while saving GPU memory. We show this for the example of reversible GNNs and build RevGNNs with different SOTA GNN operators: GAT \citep{veli2018gat}, GCN \citep{kipf2017semi}, GraphSAGE \citep{hamilton2017inductive}, and ResGEN \citep{li2020deepergcn}. We compare them to their non-reversible residual counterparts on the \emph{ogbn-arxiv} in \tblLabel \ref{table:ogbn_arxiv_ablation}. Since \emph{ogbn-arxiv} is much smaller than the \emph{ogbn-proteins}, we are able to run all experiments with full-batch training and report statistics across 10 training runs. We observe that all of the RevGNNs consistently outperform their vanilla residual counterparts with the same channel size. The RevGNNs use less memory due to reversible connections and fewer parameters due to grouping. We increase the channel size of RevGNNs to roughly match the number of parameters of the corresponding ResGNNs, thus increasing the performance gap further. For instance, the RevGCN with 28 layers and 180 channels reduces the memory footprint by more than 75\% while improving the accurracy by 0.76\%, as compared to the ResGCN with 28 layers and 128 channels. Utilizing label propagation, the RevGAT with 5 layers and 1068 channels (3 attention heads with 356 channels for each head) achieves SOTA results on \emph{ogbn-arxiv}.

\subsection{Full-batch \vs Mini-batch Training}

\begin{figure}[!t]
    \centering
    \includegraphics[trim=0cm 1cm 0cm 0cm, width=\columnwidth]{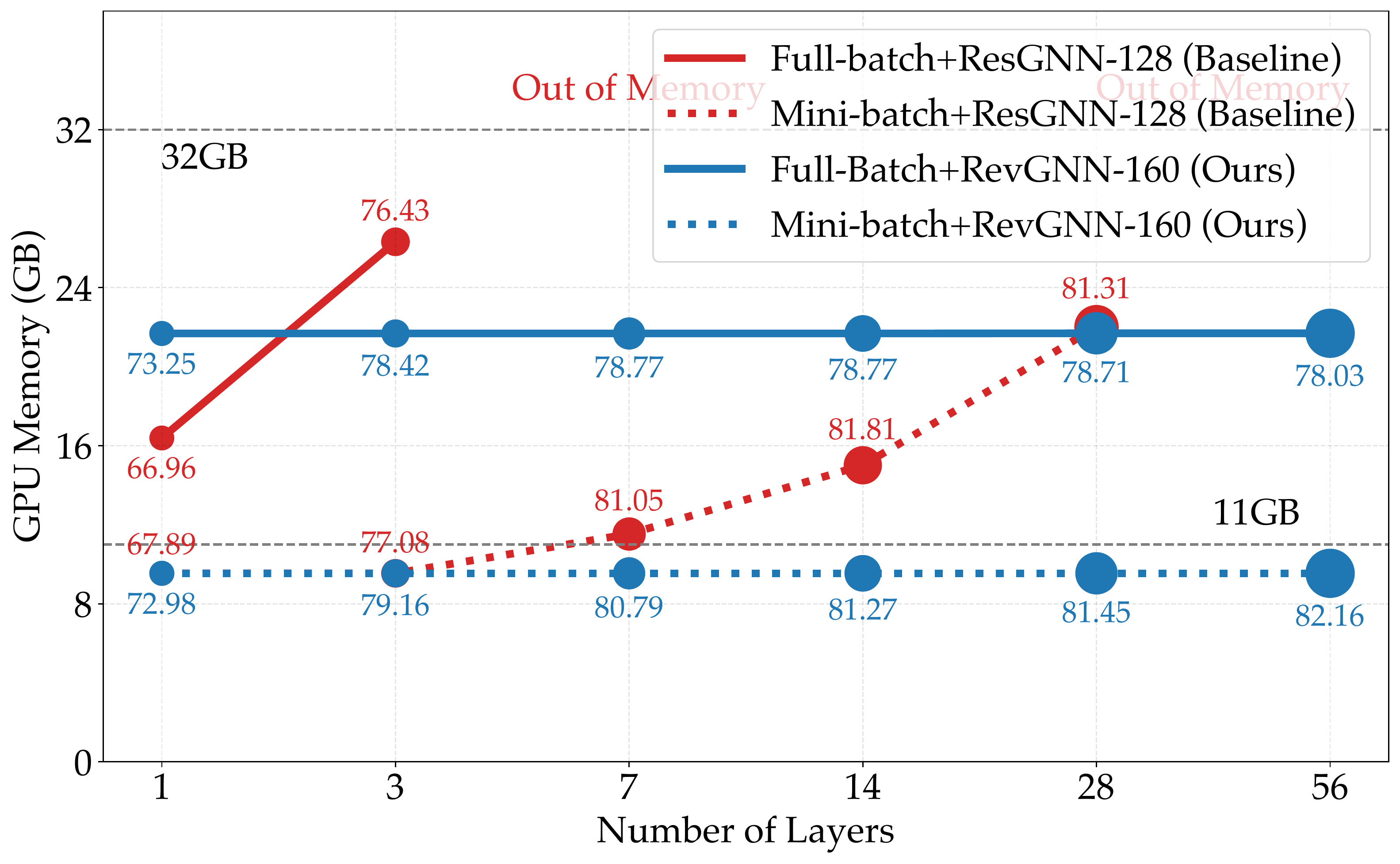}
    \caption{\textbf{GPU memory consumption \vs number of layers for ResGNN \citep{li2020deepergcn} and RevGNN with full-batch and mini-batch training on the \emph{ogbn-products}}. RevGNN uses constant memory for deeper networks with more parameters and better performance. We use a width of $128$ and $160$ for ResGNN and RevGNN, respectively,  to ensure a similar number of parameters per network depth. Datapoints are annotated with the accuracy of the model and their size is proportional to $\sqrt{p}$, where $p$ is the number of parameters.}
    \vspace{-11pt}
    \label{fig:mini_full_batch}
\end{figure}
%%%%%%%%%%%%%%

Our method is orthogonal to existing sampling-based approaches, which also reduce memory consumption. Hence, we can use our techniques in conjunction with mini-batch training to further optimize memory.
We conduct an ablation study on the \emph{ogbn-products} dataset \citep{hu2020open} with full-batch training and a simple random-clustering mini-batch training for ResGNNs and RevGNNs. The test results are reported in \figLabel \ref{fig:mini_full_batch} with full-batch inference. Compared to full-batch training, we find that mini-batch training further reduces the memory consumption of RevGNN to 44\% and improves its accuracy from 78.77\% to 82.16\%.

\begin{table*}[ht]
\vspace{-8pt}
\centering
\setlength{\tabcolsep}{5pt}
\caption{\textbf{Comparison of complexities.} $L$ is the number of layers, $D$ is the number of hidden channels, $N$ is of the number of nodes, $B$ is the batch size of nodes and $R$ is the number of sampled neighbors of each node. $K$ is the maximum Broyden iterations.}
\vspace{2pt}
\begin{tabular}{cccc}
\toprule
  \label{table:cplx}
  \centering
Method & Memory & Params & Time \\\\
\midrule
Full-batch GNN & $\mathcal{O}(LND)$ & $\mathcal{O}(LD^{2})$ & $\mathcal{O}(L\norm{A}_0D + LND^{2})$ \\
GraphSAGE & $\mathcal{O}(R^{L}BD)$ & $\mathcal{O}(LD^{2})$ & $\mathcal{O}(R^LND^{2})$\\
VR-GCN & $\mathcal{O}(LND)$ & $\mathcal{O}(LD^{2})$ & $\mathcal{O}(L\norm{A}_0D + LND^{2} + R^LND^{2})$ \\
FastGCN & $\mathcal{O}(LRBD)$ & $\mathcal{O}(LD^{2})$ & $\mathcal{O}(RLND^{2})$ \\
Cluster-GCN & $\mathcal{O}(LBD)$ & $\mathcal{O}(LD^{2})$ & $\mathcal{O}(L\norm{A}_0D + LND^{2})$ \\
GraphSAINT & $\mathcal{O}(LBD)$ & $\mathcal{O}(LD^{2})$ & $\mathcal{O}(L\norm{A}_0D + LND^{2})$ \\
Weight-tied GNN & $\mathcal{O}(LND)$ & $\mathcal{O}(D^{2})$ & $\mathcal{O}(L\norm{A}_0D + LND^{2})$ \\
\midrule
RevGNN & $\mathcal{O}(ND)$ & $\mathcal{O}(LD^{2})$ & $\mathcal{O}(L\norm{A}_0D + LND^{2})$ \\
WT-RevGNN & $\mathcal{O}(ND)$ & $\mathcal{O}(D^{2})$ & $\mathcal{O}(L\norm{A}_0D + LND^{2})$ \\
DEQ-GNN & $\mathcal{O}(ND)$ & $\mathcal{O}(D^{2})$ & $\mathcal{O}(K\norm{A}_0D + KND^{2})$ \\
\midrule
RevGNN + Subgraph Sampling & $\mathcal{O}(BD)$ & $\mathcal{O}(LD^{2})$ & $\mathcal{O}(L\norm{A}_0D + LND^{2})$\\
WT-RevGNN + Subgraph Sampling & $\mathcal{O}(BD)$ & $\mathcal{O}(D^{2})$ & $\mathcal{O}(L\norm{A}_0D + LND^{2})$ \\
DEQ-GNN + Subgraph Sampling & $\mathcal{O}(BD)$ & $\mathcal{O}(D^{2})$ & $\mathcal{O}(K\norm{A}_0D + KND^{2})$ \\
\bottomrule
\end{tabular}
\end{table*}
\subsection{Analysis of Complexities}
We have discussed the memory complexity of full-batch GNNs, GraphSAGE \citep{hamilton2017inductive}, VR-GCN \citep{chen2018stochastic}, FastGCN \citep{chen2018fastgcn}, Cluster-GCN \citep{chiang2019cluster} and GraphSAINT \citep{graphsaint-iclr20} in the related work section, and the memory complexity of our Reversible GNN, Weight-tied GNN and DEQ-GNN in the methodology section. We summarize the theoretical memory complexity in \tblLabel \ref{table:cplx}, where $L$ is the number of layers of the GNN, $D$ is the size of hidden channels, $N$ is of the number of nodes in the graph, $B$ is the batch size of nodes and $R$ is the number of sampled neighbors of each node. $K$ is the maximum Broyden iterations for equilibrium GNNs. We only discuss the memory complexity for storing intermediate node features in each layer since the memory footprint of the network parameters is negligible. All prior works suffer from memory consumption with respect to the number of layers, while the memory consumption of our methods is independent of the depth. Our methods can also be combined with mini-batch sampling methods to further reduce the memory complexity with respect to the number of nodes. We also include the parameter complexity and time complexity in \tblLabel \ref{table:cplx}. Note that although the time complexity of RevGNN is the same as for the vanilla GNNs, it has a larger constant term. This is due to the input that needs to be reconstructed during the backward pass. Both the memory complexity and parameter complexity of WT-RevGNN and DEQ-GNN is independent of $L$.

\subsection{More Ablation Studies}

\mysection{Comparison with SGC \& SIGN on Ogbn-products} 
We compare RevGNN with SGC \citep{wu2019simplifying} and SIGN \citep{sign_icml_grl2020} on ogbn-products. The test accuracies (\%) of SGC, SIGN and RevGNN on \emph{ogbn-products} are: 74.87 ± 0.25 (SGC), 77.60 ± 0.13 (SIGN) and 82.16 ± 0.15 (RevGNN) respectively.
The results of SGC and SIGN are obtained from the SIGN paper \citep{sign_icml_grl2020}. We train the RevGNN with 56 layers and 160 hidden channels in a random-clustering mini-batch training fashion. We find that RevGNN outperforms SGC and SIGN by a large margin.

\mysection{Results of RevGNN on Ogbg-ppa} 
We trained ResGNN and RevGNN with 28 layers and similar number of parameters ($\sim2.3$M) on the \emph{ogbg-ppa} dataset. Both achieve around 77\% test accuracy, but RevGNN uses only 16\% GPU memory compared to ResGNN.

\mysection{Comparison of Normalizations on Ogbg-molhiv}
We conduct an ablation study on ogbg-molhiv for comparing using BatchNorm \citep{ioffe2015batch} or GraphNorm \citep{cai2020graphnorm} for RevGNN on graph property prediction. We train RevGNN with 14 layers and 256 hidden channels. Dropout layers with a rate of $0.3$ are used. Learnable $\text{SoftMax}$ aggregation functions \citep{li2020deepergcn} are used for message aggregation. The RevGNN with GraphNorm achieves $78.62$ ROC-AUC which outperforms the RevGNN with BatchNorm ($77.82$ ROC-AUC) slightly.

\section{Conclusion and Discussion}
This work addresses a fundamental bottleneck of current deep GNNs: their high GPU memory consumption. We investigate several techniques to reduce the memory complexity with respect to network depth ($L$) from $\mathcal{O}(L)$ to $\mathcal{O}(1)$. In particular, reversible GNNs enable networks that are an order of magnitude deeper than current SOTA models. Since there is no memory cost associated with depth, the width can be increased for additional representational power. As a result, we can train over-parameterized networks that significantly outperform current models  on standard benchmarks while consuming less memory.

However, this comes at an additional cost in training time. While this cost is less than 40\%, it can extend the training time by several days on very large datasets. In addition, current GNNs are usually trained for thousands of epochs to achieve SOTA performance. Reducing the training time of GNNs is an interesting avenue for future research. Please refer to the appendix for more details about the opportunities for future work and further discussion on topics like gradient checkpointing, model parallelism, \etc~In the appendix, we also provide all the experimental details and further ablation studies.

\section*{Acknowledgments}
The authors thank Shaojie Bai, researchers at Intel ISL, the reviewers, and area chairs for their helpful suggestions. This project is partially funded by the King Abdullah University of Science and Technology (KAUST) Office of Sponsored Research under Award No. OSR-CRG2019-4033. 

\bibliography{references}

\begin{thebibliography}{77}
\providecommand{\natexlab}[1]{#1}
\providecommand{\url}[1]{\texttt{#1}}
\expandafter\ifx\csname urlstyle\endcsname\relax
  \providecommand{\doi}[1]{doi: #1}\else
  \providecommand{\doi}{doi: \begingroup \urlstyle{rm}\Url}\fi

\bibitem[Bai et~al.(2018)Bai, Kolter, and Koltun]{bai2018trellis}
Bai, S., Kolter, J.~Z., and Koltun, V.
\newblock Trellis networks for sequence modeling.
\newblock In \emph{International Conference on Learning Representations}, 2018.

\bibitem[Bai et~al.(2019)Bai, Kolter, and Koltun]{bai2019deep}
Bai, S., Kolter, J.~Z., and Koltun, V.
\newblock Deep equilibrium models.
\newblock In \emph{Advances in Neural Information Processing Systems}, 2019.

\bibitem[Bai et~al.(2020)Bai, Koltun, and Kolter]{bai2020multiscale}
Bai, S., Koltun, V., and Kolter, J.~Z.
\newblock Multiscale deep equilibrium models.
\newblock In \emph{Advances in Neural Information Processing Systems}, 2020.

\bibitem[Belkin et~al.(2019)Belkin, Hsu, Ma, and Mandal]{Belkin2019}
Belkin, M., Hsu, D., Ma, S., and Mandal, S.
\newblock Reconciling modern machine-learning practice and the classical
  bias{\textendash}variance trade-off.
\newblock \emph{Proceedings of the National Academy of Sciences}, 116\penalty0
  (32), 2019.

\bibitem[Bojchevski et~al.(2020)Bojchevski, Klicpera, Perozzi, Kapoor, Blais,
  R{\'o}zemberczki, Lukasik, and G{\"u}nnemann]{bojchevski2020scaling}
Bojchevski, A., Klicpera, J., Perozzi, B., Kapoor, A., Blais, M.,
  R{\'o}zemberczki, B., Lukasik, M., and G{\"u}nnemann, S.
\newblock Scaling graph neural networks with approximate pagerank.
\newblock In \emph{Proceedings of the 26th ACM SIGKDD International Conference
  on Knowledge Discovery \& Data Mining}, pp.\  2464--2473, 2020.

\bibitem[Brown et~al.(2020)Brown, Mann, Ryder, Subbiah, Kaplan, Dhariwal,
  Neelakantan, Shyam, Sastry, Askell, Agarwal, Herbert-Voss, Krueger, Henighan,
  Child, Ramesh, Ziegler, Wu, Winter, Hesse, Chen, Sigler, Litwin, Gray, Chess,
  Clark, Berner, McCandlish, Radford, Sutskever, and Amodei]{brown2020language}
Brown, T., Mann, B., Ryder, N., Subbiah, M., Kaplan, J.~D., Dhariwal, P.,
  Neelakantan, A., Shyam, P., Sastry, G., Askell, A., Agarwal, S.,
  Herbert-Voss, A., Krueger, G., Henighan, T., Child, R., Ramesh, A., Ziegler,
  D., Wu, J., Winter, C., Hesse, C., Chen, M., Sigler, E., Litwin, M., Gray,
  S., Chess, B., Clark, J., Berner, C., McCandlish, S., Radford, A., Sutskever,
  I., and Amodei, D.
\newblock Language models are few-shot learners.
\newblock In Larochelle, H., Ranzato, M., Hadsell, R., Balcan, M.~F., and Lin,
  H. (eds.), \emph{Advances in Neural Information Processing Systems},
  volume~33, pp.\  1877--1901, 2020.

\bibitem[Broyden(1965)]{broyden1965class}
Broyden, C.~G.
\newblock A class of methods for solving nonlinear simultaneous equations.
\newblock \emph{Mathematics of computation}, 19\penalty0 (92):\penalty0
  577--593, 1965.

\bibitem[Bruna et~al.(2014)Bruna, Zaremba, Szlam, and Lecun]{bruna2013spectral}
Bruna, J., Zaremba, W., Szlam, A., and Lecun, Y.
\newblock Spectral networks and locally connected networks on graphs.
\newblock In \emph{International Conference on Learning Representations}, 2014.

\bibitem[Cai et~al.(2020)Cai, Luo, Xu, He, Liu, and Wang]{cai2020graphnorm}
Cai, T., Luo, S., Xu, K., He, D., Liu, T.-y., and Wang, L.
\newblock Graphnorm: A principled approach to accelerating graph neural network
  training.
\newblock \emph{arXiv preprint arXiv:2009.03294}, 2020.

\bibitem[Chen et~al.(2018{\natexlab{a}})Chen, Ma, and Xiao]{chen2018fastgcn}
Chen, J., Ma, T., and Xiao, C.
\newblock Fastgcn: Fast learning with graph convolutional networks via
  importance sampling.
\newblock In \emph{International Conference on Learning Representations},
  2018{\natexlab{a}}.

\bibitem[Chen et~al.(2018{\natexlab{b}})Chen, Zhu, and
  Song]{chen2018stochastic}
Chen, J., Zhu, J., and Song, L.
\newblock Stochastic training of graph convolutional networks with variance
  reduction.
\newblock In \emph{International Conference on Machine Learning}, pp.\
  941--949, 2018{\natexlab{b}}.

\bibitem[Chen et~al.(2020)Chen, Wei, Huang, Ding, and Li]{chen2020simple}
Chen, M., Wei, Z., Huang, Z., Ding, B., and Li, Y.
\newblock Simple and deep graph convolutional networks.
\newblock In \emph{Proceedings of the 37th International Conference on Machine
  Learning}, volume 119, pp.\  1725--1735, 2020.

\bibitem[Chiang et~al.(2019)Chiang, Liu, Si, Li, Bengio, and
  Hsieh]{chiang2019cluster}
Chiang, W.-L., Liu, X., Si, S., Li, Y., Bengio, S., and Hsieh, C.-J.
\newblock Cluster-gcn: An efficient algorithm for training deep and large graph
  convolutional networks.
\newblock In \emph{Proceedings of the 25th ACM SIGKDD International Conference
  on Knowledge Discovery \& Data Mining}, pp.\  257--266, 2019.

\bibitem[Defferrard et~al.(2016)Defferrard, Bresson, and
  Vandergheynst]{defferrard2016convolutional}
Defferrard, M., Bresson, X., and Vandergheynst, P.
\newblock Convolutional neural networks on graphs with fast localized spectral
  filtering.
\newblock In \emph{Advances in neural information processing systems}, pp.\
  3844--3852, 2016.

\bibitem[Devlin et~al.(2019)Devlin, Chang, Lee, and Toutanova]{devlin2018bert}
Devlin, J., Chang, M.-W., Lee, K., and Toutanova, K.
\newblock Bert: Pre-training of deep bidirectional transformers for language
  understanding.
\newblock In \emph{Proceedings of the 2019 Conference of the North American
  Chapter of the Association for Computational Linguistics: Human Language
  Technologies}, volume~1, pp.\  4171--4186, 2019.

\bibitem[Dwivedi et~al.(2020)Dwivedi, Joshi, Laurent, Bengio, and
  Bresson]{dwivedi2020benchmarkgnns}
Dwivedi, V.~P., Joshi, C.~K., Laurent, T., Bengio, Y., and Bresson, X.
\newblock Benchmarking graph neural networks.
\newblock \emph{arXiv preprint arXiv:2003.00982}, 2020.

\bibitem[Fey \& Lenssen(2019)Fey and Lenssen]{Fey/Lenssen/2019}
Fey, M. and Lenssen, J.~E.
\newblock Fast graph representation learning with {PyTorch Geometric}.
\newblock In \emph{ICLR Workshop on Representation Learning on Graphs and
  Manifolds}, 2019.

\bibitem[Frasca et~al.(2020)Frasca, Rossi, Eynard, Chamberlain, Bronstein, and
  Monti]{sign_icml_grl2020}
Frasca, F., Rossi, E., Eynard, D., Chamberlain, B., Bronstein, M., and Monti,
  F.
\newblock Sign: Scalable inception graph neural networks.
\newblock In \emph{ICML 2020 Workshop on Graph Representation Learning and
  Beyond}, 2020.

\bibitem[Gao et~al.(2018)Gao, Wang, and Ji]{gao2018large}
Gao, H., Wang, Z., and Ji, S.
\newblock Large-scale learnable graph convolutional networks.
\newblock In \emph{Proceedings of the 24th ACM SIGKDD International Conference
  on Knowledge Discovery \& Data Mining}, pp.\  1416--1424, 2018.

\bibitem[Gilmer et~al.(2017)Gilmer, Schoenholz, Riley, Vinyals, and
  Dahl]{gilmer2017neural}
Gilmer, J., Schoenholz, S.~S., Riley, P.~F., Vinyals, O., and Dahl, G.~E.
\newblock Neural message passing for quantum chemistry.
\newblock In \emph{Proceedings of the 34th International Conference on Machine
  Learning}, 2017.

\bibitem[Gomez et~al.(2017)Gomez, Ren, Urtasun, and
  Grosse]{gomez2017reversible}
Gomez, A.~N., Ren, M., Urtasun, R., and Grosse, R.~B.
\newblock The reversible residual network: Backpropagation without storing
  activations.
\newblock In \emph{Advances in Neural Information Processing Systems}, 2017.

\bibitem[Gong et~al.(2020)Gong, Bahri, Bronstein, and
  Zafeiriou]{gong2020geometrically}
Gong, S., Bahri, M., Bronstein, M.~M., and Zafeiriou, S.
\newblock Geometrically principled connections in graph neural networks.
\newblock In \emph{Proceedings of the IEEE/CVF Conference on Computer Vision
  and Pattern Recognition}, pp.\  11415--11424, 2020.

\bibitem[Gori et~al.(2005)Gori, Monfardini, and Scarselli]{gori2005new}
Gori, M., Monfardini, G., and Scarselli, F.
\newblock A new model for learning in graph domains.
\newblock In \emph{Proceedings. 2005 IEEE International Joint Conference on
  Neural Networks, 2005.}, volume~2, pp.\  729--734. IEEE, 2005.

\bibitem[Hamilton et~al.(2017)Hamilton, Ying, and
  Leskovec]{hamilton2017inductive}
Hamilton, W., Ying, Z., and Leskovec, J.
\newblock Inductive representation learning on large graphs.
\newblock In \emph{Advances in neural information processing systems}, pp.\
  1024--1034, 2017.

\bibitem[Hasanzadeh et~al.(2020)Hasanzadeh, Hajiramezanali, Boluki, Zhou,
  Duffield, Narayanan, and Qian]{hasanzadeh2020bayesian}
Hasanzadeh, A., Hajiramezanali, E., Boluki, S., Zhou, M., Duffield, N.,
  Narayanan, K., and Qian, X.
\newblock Bayesian graph neural networks with adaptive connection sampling.
\newblock In \emph{International Conference on Machine Learning}, pp.\
  4094--4104. PMLR, 2020.

\bibitem[He et~al.(2016)He, Zhang, Ren, and Sun]{he2016deep}
He, K., Zhang, X., Ren, S., and Sun, J.
\newblock Deep residual learning for image recognition.
\newblock In \emph{Proceedings of the IEEE conference on computer vision and
  pattern recognition}, pp.\  770--778, 2016.

\bibitem[Henaff et~al.(2015)Henaff, Bruna, and LeCun]{henaff2015deep}
Henaff, M., Bruna, J., and LeCun, Y.
\newblock Deep convolutional networks on graph-structured data.
\newblock \emph{arXiv preprint arXiv:1506.05163}, 2015.

\bibitem[Hu et~al.(2020)Hu, Fey, Zitnik, Dong, Ren, Liu, Catasta, and
  Leskovec]{hu2020open}
Hu, W., Fey, M., Zitnik, M., Dong, Y., Ren, H., Liu, B., Catasta, M., and
  Leskovec, J.
\newblock Open graph benchmark: Datasets for machine learning on graphs.
\newblock In \emph{Advances in Neural Information Processing Systems},
  volume~33, pp.\  22118--22133, 2020.

\bibitem[Hu et~al.(2021)Hu, Fey, Ren, Nakata, Dong, and Leskovec]{hu2021ogblsc}
Hu, W., Fey, M., Ren, H., Nakata, M., Dong, Y., and Leskovec, J.
\newblock Ogb-lsc: A large-scale challenge for machine learning on graphs.
\newblock \emph{arXiv preprint arXiv:2103.09430}, 2021.

\bibitem[Huang et~al.(2017)Huang, Liu, Van Der~Maaten, and
  Weinberger]{huang2017densely}
Huang, G., Liu, Z., Van Der~Maaten, L., and Weinberger, K.~Q.
\newblock Densely connected convolutional networks.
\newblock In \emph{Proceedings of the IEEE conference on computer vision and
  pattern recognition}, pp.\  4700--4708, 2017.

\bibitem[Inan et~al.(2017)Inan, Khosravi, and Socher]{inan2016tying}
Inan, H., Khosravi, K., and Socher, R.
\newblock Tying word vectors and word classifiers: A loss framework for
  language modeling.
\newblock In \emph{Proceedings of the 5th International Conference on Learning
  Representations}, 2017.

\bibitem[Ioffe \& Szegedy(2015)Ioffe and Szegedy]{ioffe2015batch}
Ioffe, S. and Szegedy, C.
\newblock Batch normalization: Accelerating deep network training by reducing
  internal covariate shift.
\newblock \emph{arXiv preprint arXiv:1502.03167}, 2015.

\bibitem[Khamsi \& Kirk(2001)Khamsi and Kirk]{khamsi2001introduction}
Khamsi, M.~A. and Kirk, W.~A.
\newblock \emph{An introduction to metric spaces and fixed point theory},
  volume~53.
\newblock John Wiley \& Sons, 2001.

\bibitem[Kipf \& Welling(2017)Kipf and Welling]{kipf2017semi}
Kipf, T.~N. and Welling, M.
\newblock Semi-supervised classification with graph convolutional networks.
\newblock In \emph{International Conference on Learning Representations}, 2017.

\bibitem[Kitaev et~al.(2019)Kitaev, Kaiser, and Levskaya]{kitaev2019reformer}
Kitaev, N., Kaiser, L., and Levskaya, A.
\newblock Reformer: The efficient transformer.
\newblock In \emph{International Conference on Learning Representations}, 2019.

\bibitem[Klicpera et~al.(2019)Klicpera, Bojchevski, and
  G{\"u}nnemann]{klicpera_predict_2019}
Klicpera, J., Bojchevski, A., and G{\"u}nnemann, S.
\newblock Predict then propagate: Graph neural networks meet personalized
  pagerank.
\newblock In \emph{International Conference on Learning Representations}, 2019.

\bibitem[Krizhevsky et~al.(2012)Krizhevsky, Sutskever, and
  Hinton]{krizhevsky2012imagenet}
Krizhevsky, A., Sutskever, I., and Hinton, G.~E.
\newblock Imagenet classification with deep convolutional neural networks.
\newblock \emph{Advances in neural information processing systems},
  25:\penalty0 1097--1105, 2012.

\bibitem[Leemput et~al.(2019)Leemput, Teuwen, Ginneken, and
  Manniesing]{vandeLeemput2019MemCNN}
Leemput, S. C.~v., Teuwen, J., Ginneken, B.~v., and Manniesing, R.
\newblock Memcnn: A python/pytorch package for creating memory-efficient
  invertible neural networks.
\newblock \emph{Journal of Open Source Software}, 4\penalty0 (39):\penalty0
  1576, 7 2019.
\newblock ISSN 2475-9066.

\bibitem[Lepikhin et~al.(2021)Lepikhin, Lee, Xu, Chen, Firat, Huang, Krikun,
  Shazeer, and Chen]{lepikhin2020gshard}
Lepikhin, D., Lee, H., Xu, Y., Chen, D., Firat, O., Huang, Y., Krikun, M.,
  Shazeer, N., and Chen, Z.
\newblock Gshard: Scaling giant models with conditional computation and
  automatic sharding.
\newblock In \emph{International Conference on Learning Representations}, 2021.

\bibitem[Levie et~al.(2018)Levie, Monti, Bresson, and
  Bronstein]{levie2018cayleynets}
Levie, R., Monti, F., Bresson, X., and Bronstein, M.~M.
\newblock Cayleynets: Graph convolutional neural networks with complex rational
  spectral filters.
\newblock \emph{IEEE Transactions on Signal Processing}, 67\penalty0
  (1):\penalty0 97--109, 2018.

\bibitem[Li et~al.(2019)Li, Müller, Thabet, and Ghanem]{li2019deepgcns}
Li, G., Müller, M., Thabet, A., and Ghanem, B.
\newblock Deepgcns: Can gcns go as deep as cnns?
\newblock In \emph{The IEEE International Conference on Computer Vision}, 2019.

\bibitem[Li et~al.(2020)Li, Xiong, Thabet, and Ghanem]{li2020deepergcn}
Li, G., Xiong, C., Thabet, A., and Ghanem, B.
\newblock Deepergcn: All you need to train deeper gcns.
\newblock \emph{arXiv preprint arXiv:2006.07739}, 2020.

\bibitem[Li et~al.(2021)Li, M{\"u}ller, Qian, Perez, Abualshour, Thabet, and
  Ghanem]{li2021deepgcns_pami}
Li, G., M{\"u}ller, M., Qian, G., Perez, I. C.~D., Abualshour, A., Thabet,
  A.~K., and Ghanem, B.
\newblock Deepgcns: Making gcns go as deep as cnns.
\newblock \emph{IEEE Transactions on Pattern Analysis and Machine
  Intelligence}, 2021.

\bibitem[{Li} et~al.(2018){Li}, {Han}, and {Wu}]{li2018deeper}
{Li}, Q., {Han}, Z., and {Wu}, X.
\newblock {Deeper Insights into Graph Convolutional Networks for
  Semi-Supervised Learning}.
\newblock In \emph{The Thirty-Second AAAI Conference on Artificial
  Intelligence}, 2018.

\bibitem[Li et~al.(2018)Li, Wang, Zhu, and Huang]{li2018adaptive}
Li, R., Wang, S., Zhu, F., and Huang, J.
\newblock Adaptive graph convolutional neural networks.
\newblock In \emph{Thirty-second AAAI conference on artificial intelligence},
  2018.

\bibitem[Liu et~al.(2019)Liu, Kumar, Ba, Kiros, and Swersky]{liu2019gnf}
Liu, J., Kumar, A., Ba, J., Kiros, J., and Swersky, K.
\newblock Graph normalizing flows.
\newblock \emph{Advances in Neural Information Processing Systems}, 2019.

\bibitem[Liu et~al.(2020)Liu, Gao, and Ji]{liu2020towards}
Liu, M., Gao, H., and Ji, S.
\newblock Towards deeper graph neural networks.
\newblock In \emph{Proceedings of the 26th ACM SIGKDD International Conference
  on Knowledge Discovery \& Data Mining}, pp.\  338--348, 2020.

\bibitem[Monti et~al.(2017)Monti, Boscaini, Masci, Rodola, Svoboda, and
  Bronstein]{monti2017geometric}
Monti, F., Boscaini, D., Masci, J., Rodola, E., Svoboda, J., and Bronstein,
  M.~M.
\newblock Geometric deep learning on graphs and manifolds using mixture model
  cnns.
\newblock In \emph{Proceedings of the IEEE Conference on Computer Vision and
  Pattern Recognition}, pp.\  5115--5124, 2017.

\bibitem[Neyshabur et~al.(2019)Neyshabur, Li, Bhojanapalli, LeCun, and
  Srebro]{neyshabur2018towards}
Neyshabur, B., Li, Z., Bhojanapalli, S., LeCun, Y., and Srebro, N.
\newblock The role of over-parametrization in generalization of neural
  networks.
\newblock In \emph{International Conference on Learning Representations}, 2019.

\bibitem[Niepert et~al.(2016)Niepert, Ahmed, and Kutzkov]{niepert2016learning}
Niepert, M., Ahmed, M., and Kutzkov, K.
\newblock Learning convolutional neural networks for graphs.
\newblock In \emph{International conference on machine learning}, pp.\
  2014--2023, 2016.

\bibitem[Oono \& Suzuki(2019)Oono and Suzuki]{oono2019graph}
Oono, K. and Suzuki, T.
\newblock Graph neural networks exponentially lose expressive power for node
  classification.
\newblock In \emph{International Conference on Learning Representations}, 2019.

\bibitem[Paszke et~al.(2019)Paszke, Gross, Massa, Lerer, Bradbury, Chanan,
  Killeen, Lin, Gimelshein, Antiga, Desmaison, Kopf, Yang, DeVito, Raison,
  Tejani, Chilamkurthy, Steiner, Fang, Bai, and Chintala]{pytorch2019}
Paszke, A., Gross, S., Massa, F., Lerer, A., Bradbury, J., Chanan, G., Killeen,
  T., Lin, Z., Gimelshein, N., Antiga, L., Desmaison, A., Kopf, A., Yang, E.,
  DeVito, Z., Raison, M., Tejani, A., Chilamkurthy, S., Steiner, B., Fang, L.,
  Bai, J., and Chintala, S.
\newblock Pytorch: An imperative style, high-performance deep learning library.
\newblock In Wallach, H., Larochelle, H., Beygelzimer, A., d\textquotesingle
  Alch\'{e}-Buc, F., Fox, E., and Garnett, R. (eds.), \emph{Advances in Neural
  Information Processing Systems 32}, pp.\  8024--8035. Curran Associates,
  Inc., 2019.

\bibitem[Press \& Wolf(2017)Press and Wolf]{press2017using}
Press, O. and Wolf, L.
\newblock Using the output embedding to improve language models.
\newblock In \emph{Proceedings of the 15th Conference of the European Chapter
  of the Association for Computational Linguistics: Volume 2, Short Papers},
  pp.\  157--163, 2017.

\bibitem[Radford et~al.()Radford, Narasimhan, Salimans, and
  Sutskever]{radford2018improving}
Radford, A., Narasimhan, K., Salimans, T., and Sutskever, I.
\newblock Improving language understanding by generative pre-training.

\bibitem[Rasley et~al.(2020)Rasley, Rajbhandari, Ruwase, and
  He]{rasley2020deepspeed}
Rasley, J., Rajbhandari, S., Ruwase, O., and He, Y.
\newblock Deepspeed: System optimizations enable training deep learning models
  with over 100 billion parameters.
\newblock In \emph{Proceedings of the 26th ACM SIGKDD International Conference
  on Knowledge Discovery \& Data Mining}, pp.\  3505--3506, 2020.

\bibitem[Rong et~al.(2020)Rong, Huang, Xu, and Huang]{rong2020dropedge}
Rong, Y., Huang, W., Xu, T., and Huang, J.
\newblock Dropedge: Towards deep graph convolutional networks on node
  classification.
\newblock In \emph{International Conference on Learning Representations}, 2020.

\bibitem[Scarselli et~al.(2008)Scarselli, Gori, Tsoi, Hagenbuchner, and
  Monfardini]{scarselli2008graph}
Scarselli, F., Gori, M., Tsoi, A.~C., Hagenbuchner, M., and Monfardini, G.
\newblock The graph neural network model.
\newblock \emph{IEEE transactions on neural networks}, 20\penalty0
  (1):\penalty0 61--80, 2008.

\bibitem[Senior et~al.(2020)Senior, Evans, Jumper, Kirkpatrick, Sifre, Green,
  Qin, {\v{Z}}{\'\i}dek, Nelson, Bridgland, et~al.]{senior2020improved}
Senior, A.~W., Evans, R., Jumper, J., Kirkpatrick, J., Sifre, L., Green, T.,
  Qin, C., {\v{Z}}{\'\i}dek, A., Nelson, A.~W., Bridgland, A., et~al.
\newblock Improved protein structure prediction using potentials from deep
  learning.
\newblock \emph{Nature}, 577\penalty0 (7792):\penalty0 706--710, 2020.

\bibitem[Shchur et~al.(2018)Shchur, Mumme, Bojchevski, and
  G{\"u}nnemann]{shchur2018pitfalls}
Shchur, O., Mumme, M., Bojchevski, A., and G{\"u}nnemann, S.
\newblock Pitfalls of graph neural network evaluation.
\newblock \emph{Relational Representation Learning Workshop, NeurIPS}, 2018.

\bibitem[Shi et~al.(2020)Shi, Huang, Wang, Zhong, Feng, and Sun]{shi2020masked}
Shi, Y., Huang, Z., Wang, W., Zhong, H., Feng, S., and Sun, Y.
\newblock Masked label prediction: Unified message passing model for
  semi-supervised classification.
\newblock \emph{arXiv preprint arXiv:2009.03509}, 2020.

\bibitem[Veličković et~al.(2018)Veličković, Cucurull, Casanova, Romero,
  Liò, and Bengio]{veli2018gat}
Veličković, P., Cucurull, G., Casanova, A., Romero, A., Liò, P., and Bengio,
  Y.
\newblock Graph attention networks.
\newblock In \emph{International Conference on Learning Representations}, 2018.

\bibitem[Wang et~al.(2020)Wang, Shen, Huang, Wu, Dong, and
  Kanakia]{wang2020microsoft}
Wang, K., Shen, I., Huang, C., Wu, C.-H., Dong, Y., and Kanakia, A.
\newblock Microsoft academic graph: when experts are not enough.
\newblock \emph{Quantitative Science Studies}, 1\penalty0 (1):\penalty0
  396--413, February 2020.

\bibitem[Wang et~al.(2019{\natexlab{a}})Wang, Zheng, Ye, Gan, Li, Song, Zhou,
  Ma, Yu, Gai, Xiao, He, Karypis, Li, and Zhang]{wang2019dgl}
Wang, M., Zheng, D., Ye, Z., Gan, Q., Li, M., Song, X., Zhou, J., Ma, C., Yu,
  L., Gai, Y., Xiao, T., He, T., Karypis, G., Li, J., and Zhang, Z.
\newblock Deep graph library: A graph-centric, highly-performant package for
  graph neural networks.
\newblock \emph{arXiv preprint arXiv:1909.01315}, 2019{\natexlab{a}}.

\bibitem[Wang et~al.(2019{\natexlab{b}})Wang, Sun, Liu, Sarma, Bronstein, and
  Solomon]{wang2019dynamic}
Wang, Y., Sun, Y., Liu, Z., Sarma, S.~E., Bronstein, M.~M., and Solomon, J.~M.
\newblock Dynamic graph cnn for learning on point clouds.
\newblock \emph{ACM Transactions on Graphics (TOG)}, 38\penalty0 (5):\penalty0
  1--12, 2019{\natexlab{b}}.

\bibitem[Wang et~al.(2021)Wang, Jin, Zhang, Yu, Zhang, and Wipf]{wang2021bag}
Wang, Y., Jin, J., Zhang, W., Yu, Y., Zhang, Z., and Wipf, D.
\newblock Bag of tricks for node classification with graph neural networks.
\newblock \emph{arXiv preprint arXiv:2103.13355}, 2021.

\bibitem[Wu et~al.(2019)Wu, Souza, Zhang, Fifty, Yu, and
  Weinberger]{wu2019simplifying}
Wu, F., Souza, A., Zhang, T., Fifty, C., Yu, T., and Weinberger, K.
\newblock Simplifying graph convolutional networks.
\newblock In \emph{International conference on machine learning}, pp.\
  6861--6871. PMLR, 2019.

\bibitem[Xie et~al.(2017)Xie, Girshick, Doll{\'a}r, Tu, and
  He]{xie2017aggregated}
Xie, S., Girshick, R., Doll{\'a}r, P., Tu, Z., and He, K.
\newblock Aggregated residual transformations for deep neural networks.
\newblock In \emph{Proceedings of the IEEE conference on computer vision and
  pattern recognition}, pp.\  1492--1500, 2017.

\bibitem[Xu et~al.(2018)Xu, Li, Tian, Sonobe, Kawarabayashi, and
  Jegelka]{xu2018jknet}
Xu, K., Li, C., Tian, Y., Sonobe, T., Kawarabayashi, K., and Jegelka, S.
\newblock Representation learning on graphs with jumping knowledge networks.
\newblock In \emph{Proceedings of the 35th International Conference on Machine
  Learning}, 2018.

\bibitem[Xu et~al.(2019)Xu, Hu, Leskovec, and Jegelka]{xu2018powerful}
Xu, K., Hu, W., Leskovec, J., and Jegelka, S.
\newblock How powerful are graph neural networks?
\newblock In \emph{International Conference on Learning Representations}, 2019.

\bibitem[Xu et~al.(2021)Xu, Zhang, Jegelka, and Kawaguchi]{xu2021optimization}
Xu, K., Zhang, M., Jegelka, S., and Kawaguchi, K.
\newblock Optimization of graph neural networks: Implicit acceleration by skip
  connections and more depth.
\newblock \emph{arXiv preprint arXiv:2105.04550}, 2021.

\bibitem[Yang et~al.(2016)Yang, Cohen, and Salakhudinov]{yang2016revisiting}
Yang, Z., Cohen, W., and Salakhudinov, R.
\newblock Revisiting semi-supervised learning with graph embeddings.
\newblock In \emph{International conference on machine learning}, pp.\  40--48.
  PMLR, 2016.

\bibitem[Yu \& Koltun(2016)Yu and Koltun]{YuKoltun2016}
Yu, F. and Koltun, V.
\newblock Multi-scale context aggregation by dilated convolutions.
\newblock In \emph{International Conference on Learning Representations}, 2016.

\bibitem[Zeng et~al.(2020)Zeng, Zhou, Srivastava, Kannan, and
  Prasanna]{graphsaint-iclr20}
Zeng, H., Zhou, H., Srivastava, A., Kannan, R., and Prasanna, V.
\newblock {GraphSAINT}: Graph sampling based inductive learning method.
\newblock In \emph{International Conference on Learning Representations}, 2020.

\bibitem[Zhang et~al.(2019)Zhang, Song, Gao, Chen, Bao, and Ma]{Zhang2019be}
Zhang, L., Song, J., Gao, A., Chen, J., Bao, C., and Ma, K.
\newblock Be your own teacher: Improve the performance of convolutional neural
  networks via self distillation.
\newblock In \emph{Proceedings of the IEEE/CVF International Conference on
  Computer Vision}, October 2019.

\bibitem[Zhao \& Akoglu(2019)Zhao and Akoglu]{zhao2019pairnorm}
Zhao, L. and Akoglu, L.
\newblock Pairnorm: Tackling oversmoothing in gnns.
\newblock In \emph{International Conference on Learning Representations}, 2019.

\bibitem[Zhou et~al.(2020)Zhou, Huang, Li, Zha, Chen, and Hu]{zhou2020towards}
Zhou, K., Huang, X., Li, Y., Zha, D., Chen, R., and Hu, X.
\newblock Towards deeper graph neural networks with differentiable group
  normalization.
\newblock \emph{Advances in Neural Information Processing Systems}, 33, 2020.

\bibitem[Zitnik \& Leskovec(2017)Zitnik and Leskovec]{zitnik2017predicting}
Zitnik, M. and Leskovec, J.
\newblock Predicting multicellular function through multi-layer tissue
  networks.
\newblock \emph{Bioinformatics}, 33\penalty0 (14):\penalty0 i190--i198, 2017.

\end{thebibliography}
\bibliographystyle{icml2021}

\clearpage
\appendix
\section{Grouping of RevGNN}
Grouped convolution is an effective way to reduce the parameter complexity in CNNs. We provide an ablation study to show how grouping reduces the number of parameters of RevGNNs. We conduct experiments on the \textit{ogbn-proteins} dataset with different group sizes and report the results in \tblLabel \ref{table:group}. The number of hidden channels for all of these models is set to 224. We find that a larger group size reduces the number of parameters. As the group size increases from 2 to 4, the number of parameters reduces by more than 30\%. The performance of models with 3 to 56 layers decreases slightly. The 112-layer networks achieve the same performance while the model with group size 4 uses only around 67\% parameters compared to the model with group size 2. However, we observe that the GPU memory usage increases from 7.30 GB to 11.05 GB as the group size increases from 2 to 4 with our current implementation. We conjecture that this is due to our inefficient implementation. Optimizing our code for larger group sizes and conducting a more rigorous analysis is an interesting avenue for future investigation.

\begin{table}[ht]
\vspace{-8pt}
\centering
\setlength{\tabcolsep}{5pt}
\caption{\textbf{Ablation of the group size of group reversible GNNs on the \textit{ogbn-protein} dataset.} $L$ is the number of layers. The number of hidden channels is 224 for all the models.}
\vspace{2pt}
\begin{tabular}{c|cc|cc}
\toprule
  \label{table:group}
  \centering
  &  \multicolumn{2}{c|}{Group=2} & \multicolumn{2}{c}{Group=4} \\
\midrule
\#L & Params & ROC-AUC $\uparrow$ & Params & ROC-AUC $\uparrow$\\
\midrule
3 & 490k & 85.09 & 339k & 84.86 \\
7 & 1.1M & 85.68 & 750k & 85.25 \\
14 & 2.2M & 86.62 & 1.5M & 85.79 \\
28 & 4.3M & 86.68 & 2.9M & 86.30 \\
56 & 8.6M & 86.90 & 5.8M & 86.76 \\
112 & 17.2M & 87.02 & 11.5M & 87.09 \\
\bottomrule
\end{tabular}
\end{table}

\section{Experimental Details and More Ablations}

\subsection{Datasets and Frameworks}
We conduct experiments on three OGB datasets \citep{hu2020open} including \textit{ogbn-proteins}, \textit{ogbn-arxiv} and \textit{ogbn-products}. We follow the standard data splits and evaluation protocol of OGB 1.2.4. Please refer to the OGB website\footnote{https://ogb.stanford.edu/} for more details. Our code implementation relies on the deep learning framework Pytorch 1.6.0. We use Pytorch Geometric 1.6.1 \citep{Fey/Lenssen/2019} for all experiments except for the experiments with GATs where we use DGL 0.5.3 \citep{wang2019dgl}. The reversible module is implemented based on MemCNN \citep{vandeLeemput2019MemCNN}. The deep equilibrium module is implemented based on DEQ \citep{bai2019deep}.

\subsection{Hyperparameters and Experimental Settings}
We describe all important hyperparameters and training settings that were not mentioned in the main paper for reproducibility. The settings are slightly different for each dataset.

\mysection{Ogbn-proteins} The node features are initialized through aggregating connected edge features by a \emph{sum} aggregation at the first layer. We use random partitioning for mini-batch training. The number of partitions is set to 10 for training and 5 for validation for all the ablated models. One subgraph is sampled at each SGD step. One layer normalization is used in the GNN block. Dropout with a rate of $0.1$ is used for each layer. We use \emph{max} as the message aggregator. Each model is trained for $2000$ epochs using the Adam optimizer with a learning rate of $0.001$.

\mysection{Ablations on Ogbn-proteins}
A detailed comparison of ResGNN, Weight-tied ResGNN, DEQ-GNN, RevGNN and Weight-tied RevGNN is shown in \tblLabel \ref{table:ogbn_proteins_ablation}. Except for the DEQ-GNN, all the other models have an explicit depth of 112 layers. The reversible connections reduce the memory consumption significantly and enable training of wider RevGNNs. A 112-layer RevGNN achieves the best performance (87.02 ROC-AUC) among the compared models. DEQ-GNN with 64 channels and WT-RevGNN with 80 channels have a similar number of parameters and memory consumption and also perform similarly. However, training DEQ-GNN is significantly faster than training WT-RevGNN (1.3 days \vs 2 days).

\begin{table}[t]
% \vspace{-8pt}
\centering
\setlength{\tabcolsep}{2pt}
\caption{\textbf{Results on the \emph{ogbn-proteins} dataset for various 112-layer networks.} Note that DEQ-GNN always has only a single layer that approximates an infinitely deep network. Each network is trained on one V100 GPU with 32GB of memory. The column \emph{Mem} reports the GPU memory in GB, \emph{Params} reports the number of model parameters, and \emph{Time} reports the training time in days. \textit{Baselines} are in italic.}
\vspace{2pt}
\begin{tabular}{cccccc}
\toprule
  \label{table:ogbn_proteins_ablation}
  \centering
Model & \#Ch & ROC-AUC $\uparrow$  & Mem $\downarrow$ & Params & Time $\downarrow$ \\
\midrule
\textit{ResGNN} & 64 & 85.94 & 27.1 
& 2.37M %2,374,456 
& 1.3 \\
\textit{ResGNN} & 224 & - & OOM 
& 28.4M %28,380,536 
& - \\
\midrule
\textit{WT-ResGNN} & 64 & 83.30 & 27.4
& 51.2k %51,256 
& 1.2 \\
\textit{WT-ResGNN} & 224 & - & OOM  
& 537k %537,336 
& - \\
\midrule
DEQ-GNN & 64 & 83.17 & 2.22 
& 51.3k %51,256 
& 1.3 \\
DEQ-GNN & 224 & 85.84 & 7.60 
& 537k %537,336 
& 2.9 \\
\midrule
RevGNN & 64 & 85.48 & 2.09
& 1.46M %1,457,080
& 1.8 \\
RevGNN & 80 & 85.97 & 2.56 
& 2.25M %2,251,384 
& 2.2 \\
RevGNN & 224 & 87.02 & 7.30
& 17.1M %17,141,560 
& 4.9 \\
\midrule
WT-RevGNN & 64 & 82.89 & 1.60 
& 35.0k %35,000 
& 1.7 \\
WT-RevGNN & 80 & 83.46 & 2.08 
& 51.4k %51,384 
& 2.0 \\
WT-RevGNN & 224 & 85.28 & 5.55 
& 337k %337,080 
& 4.8 \\
\bottomrule
\end{tabular}
\end{table}

\mysection{Multi-view Inference on Ogbn-proteins}
To further improve the evaluation results, we propose multi-view inference which reduces the negative effects of random partitioning and noisy neighbors. During different inference passes, each vertex will see a different set of neighbors. We refer to this as multi-view inference and implement it by partitioning the graphs into different subgraphs in each inference pass. 
In \tblLabel \ref{table:multi_view}, we find that performing inference with more views yields better results. We observe a substantial improvement with increasing number of views for both the RevGNN-Deep and RevGNN-Wide models. The results increase by about $0.4\%$ in terms of ROC-AUC going from $1$ view to $10$ views. We also observe that a smaller number of partitions is favorable for evaluation. To reduce memory cost, automatic mixed precision\footnote{https://developer.nvidia.com/automatic-mixed-precision} by NVIDIA is used for inference.

\begin{table}[t]
% \vspace{-8pt}
\centering
\setlength{\tabcolsep}{2pt}
\caption{\textbf{Ablations for multi-view inference with RevGNN-Deep and RevGNN-Wide on the \emph{ogbn-proteins} dataset.} \emph{L}, \emph{Ch}, \emph{Views} and \emph{Parts} denote the numbers of layers, channels, views and parts respectively. Doing inference with more views and less parts is favorable.}
\vspace{2pt}
\begin{tabular}{cccccc}
\toprule
  \label{table:multi_view}
  \centering
Model & \#L & \#Ch & \#Views & \#Parts & ROC-AUC $\uparrow$ \\
\midrule
RevGNN-Deep & 1001 & 80 & 1 & 3 &
87.29 \textcolor{gray}{\small{± 0.16}}\\
RevGNN-Deep & 1001 & 80 & 5 & 3 &
87.68 \textcolor{gray}{\small{± 0.13}}\\
RevGNN-Deep & 1001 & 80 & 10 & 3 &
\textbf{87.74} \textcolor{gray}{\small{± 0.13}}\\
\midrule
RevGNN-Wide & 448 & 224 & 1 & 3
& 87.84 \textcolor{gray}{\small{± 0.21}}\\
RevGNN-Wide & 448 & 224 & 5 & 3
& 88.20 \textcolor{gray}{\small{± 0.16}}\\
RevGNN-Wide & 448 & 224 & 10 & 3
& \textbf{88.24} \textcolor{gray}{\small{± 0.15}}\\
\midrule
RevGNN-Wide & 448 & 224 & 1 & 3
& \textbf{87.84} \textcolor{gray}{\small{± 0.21}}\\
RevGNN-Wide & 448 & 224 & 1 & 5
& 87.62 \textcolor{gray}{\small{± 0.18}}\\
RevGNN-Wide & 448 & 224 & 1 & 10
& 87.23 \textcolor{gray}{\small{± 0.22}}\\
\bottomrule
\end{tabular}
\end{table}

\mysection{Ogbn-arxiv}
The directed graph is converted into an undirected graph and self-loops are added. We use the full-batch setting for both training and testing. For the GCN \citep{kipf2017semi}, SAGE \citep{hamilton2017inductive} and GEN \citep{li2020deepergcn} models, batch normalization and dropout with a rate of $0.5$ is applied to each layer and the Adam optimizer with a learning rate of $0.001$ is used to train the models for $2000$ epochs. The GAT-based \citep{veli2018gat} models are implemented based on the OGB leaderboard submission \emph{GAT + norm. adj. + label reuse}\footnote{\small{https://github.com/Espylapiza/dgl/tree/master/examples \\ /pytorch/ogb/ogbn-arxiv}} \citep{wang2021bag}. The RevGAT models with self-knowledge distillation are implemented based on the submission \emph{GAT + label reuse + self KD}\footnote{ https://github.com/ShunliRen/dgl/tree/master/examples \\
/pytorch/ogb/ogbn-arxiv}. The teacher models and student models have the same architecture. A knowledge distillation loss is added to the student model to minimize the Kullback–Leibler divergence between the teacher's predictions and the student's predictions during training. Please refer to the Github repositories for more details about the implementation.

\mysection{Ogbn-products}
Self-loops are added to the graph. We compare RevGNNs with full-batch training and mini-batch training. For mini-batch training, the graph is randomly partitioned into $10$ subgraphs and one subgraph is sampled at each SGD step. We use full-batch testing in both scenarios. Batch normalization and dropout with a rate of $0.5$ are used for each GNN block. The model is trained using the Adam optimizer with a learning rate of $0.001$ for $1000$ epochs.

\subsection{GPU Memory Measurement}
In all the experiments, the GPU memory usage is measured as the peak GPU memory during the first training epoch. Note that the measured GPU memory is larger than the GPU memory for storing node features due to the intermediate computation and network parameters. We consider the peak GPU memory usage as a practical metric since it is the bottleneck for training neural networks. As is common practice, we use \texttt{torch.cuda.max\_memory\_allocated()} for the memory measurement. However, note that the measured peak GPU memory obtained using \texttt{torch.cuda.max\_memory\_allocated()} is usually smaller than the actual peak GPU memory obtained with NVIDIA-SMI.

\subsection{Correlation of Model Predictions}
We perform a correlation analysis on model predictions of RevGNN, Weight-tied RevGNN and DEQ-GNN. The pearson correlations of RevGNN with 1000 layers, WT-RevGNN-224 with 7 layers and DEQ-GNN-224 with 56 iterations are: 0.8571 (RevGNN \vs WT-RevGNN), 0.8565 (RevGNN \vs DEQ-GNN) and 0.8948 (WT-RevGNN \vs DEQ-GNN).

\section{More Discussion}

\subsection{Gradient Checkpointing and Model Parallelism}
By saving checkpoints every $\sqrt{L}$ steps, gradient checkpointing can achieve a memory complexity of $\mathcal{O}(\sqrt{L}ND)$, which is still higher than the memory complexity $\mathcal{O}(ND)$ of RevGNN. For 112-layer models with 64 hidden channels, ResGNN with gradient checkpointing consumes 2.5X the memory compared to RevGNN (5.22 G \vs 2.09 G) while reaching similar performance on \emph{ogbn-proteins} with similar training time.
Model parallelism is orthogonal to our approach. It would be interesting to investigate model parallelism to make RevGNN even wider with multiple GPUs.

\subsection{Going Deeper and Datasets}
In our experiments, we find that going deeper is very effective on \emph{ogbn-proteins}. It would probably be beneficial to pre-train overparameterized GNNs on larger-scale protein datasets and then apply the pre-trained models to scientific applications such as drug discovery, protein structure prediction and gene-disease associations. For the other datasets such as \emph{ogbn-products} and \emph{ogbn-arxiv}, we observe less improvement when going very deep. It is still unclear what kind of datasets benefit more from depth and overparameterization. Investigating the relationship between overparameterization and factors such as dataset size, graph modality and graph learning task is an important direction of future work to better understand when overparameterized models are beneficial. We also anticipate that overparameterized GNNs will be a promising solution to even larger datasets such as OGB-LSC \citep{hu2021ogblsc}.

\end{document}